\documentclass[final]{cvpr}

\usepackage{times}
\usepackage{epsfig}
\usepackage{graphicx}
\usepackage{amsmath}
\usepackage{amssymb}

\DeclareMathOperator*{\argmax}{argmax} 
\usepackage{tabularx}
\usepackage{multirow}
\usepackage{multicol}

\usepackage{subfigure}

\usepackage{booktabs}
\usepackage{caption}

\usepackage{cuted}
\usepackage{capt-of}

\usepackage{rotating} % package for begin{sideway} which is for rotating words in figures
\usepackage{appendix}

\usepackage[pagebackref=true,breaklinks=true,colorlinks,bookmarks=false]{hyperref}

\def\cvprPaperID{916} % *** Enter the CVPR Paper ID here
\def\confYear{CVPR 2021}

\begin{document}

\title{Self-Supervised Visibility Learning for Novel View Synthesis}

\author{Yujiao Shi \textsuperscript{\rm 1},~~
Hongdong Li \textsuperscript{\rm 1},~~
Xin Yu \textsuperscript{\rm 2}
\\
\textsuperscript{\rm 1}Australian National University and ACRV~~
% \textsuperscript{\rm 2}Australian Centre for Robotic Vision~~
\textsuperscript{\rm 2}University of Technology Sydney\\
{\tt\small yujiao.shi@anu.edu.au, hongdong.li@anu.edu.au, xin.yu@uts.edu.au}
}
\maketitle

\pagestyle{empty}
\thispagestyle{empty}

\begin{strip}
\setlength{\abovecaptionskip}{0pt}
    \setlength{\belowcaptionskip}{0pt}
\centering
\includegraphics[width=0.88\linewidth]{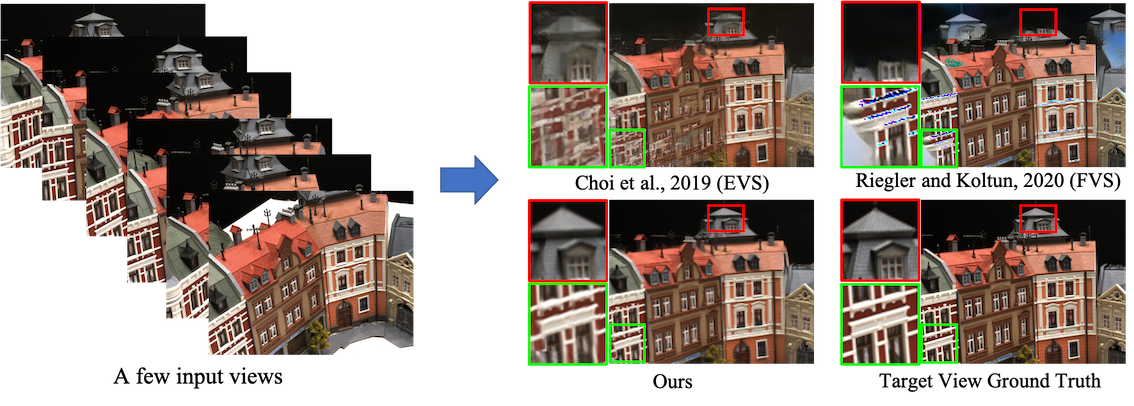}
\captionof{figure}{\footnotesize Given a few sparse and unstructured input multi-view images, our goal is to synthesize a novel view from a given target camera pose. 
Our method estimates target-view depth and source-view visibility in an end-to-end self-supervised manner. Compared with the previous state-of-the-art, such as Choi \etal~\cite{choi2019extreme} and Riegler and Koltun~\cite{riegler2020free}, our method produces superior novel view images of higher quality and with finer details, better conform to the ground-truth. 
}
\label{fig:teaser} 
\end{strip}

%%%%%%%%% ABSTRACT
\begin{abstract}

% ============== V5 =====================
We address the problem of novel view synthesis (NVS) from a few sparse source view images. 
Conventional image-based rendering methods estimate scene geometry and synthesize novel views in two separate steps. 
However, erroneous geometry estimation will decrease NVS performance as view synthesis highly depends on the quality of estimated scene geometry.
In this paper, we propose an end-to-end NVS framework to eliminate the error propagation issue. 
To be specific, we construct a volume under the target view and 
design a source-view visibility estimation (SVE) module to determine the visibility of the target-view voxels in each source view. 
Next, we aggregate the visibility of all source views to achieve a consensus volume.
Each voxel in the consensus volume indicates a surface existence probability.
Then, we present a soft ray-casting (SRC) mechanism to find the most front surface in the target view (\ie, depth). Specifically, our SRC traverses the consensus volume along viewing rays and then estimates a depth probability distribution. 
We then warp and aggregate source view pixels to synthesize a novel view based on the estimated source-view visibility and target-view depth. 
At last, our network is trained in an end-to-end self-supervised fashion, thus significantly alleviating error accumulation in view synthesis.  
Experimental results demonstrate that our method generates novel views in higher quality compared to the state-of-the-art.

\end{abstract}

%%%%%%%%% BODY TEXT
\section{Introduction}

Suppose after taking a few snapshots of a famous sculpture, we wish to look at the sculpture from some other different viewpoints.
This task would require us to generate novel-view images from the captured ones and is generally referred to as ``NVS''. 
However, compared with previous solutions, our setting is more challenging, because the number of available real views is very limited, and the underlying 3D geometry is not available.  Moreover, the occlusion along target viewing rays and the visibility of target pixels in source views are hard to infer.

% Several works\cite{mildenhall2020nerf, zhang2020nerf, sitzmann2019scene, niemeyer2020differentiable, sitzmann2019deepvoxels, thies2019deferred} propose to encode geometry and appearance of a scene from 2D observations by a deep network. 
% Although impressive performance has been demonstrated, they mostly require densely covered input images to remember all the details of a scene, \eg, NeRF~\cite{mildenhall2020nerf} takes more than 20 input views on forward facing images~\cite{mildenhall2019local} and $491$ input images on the DeepVoxel dataset~\cite{sitzmann2019deepvoxels}; NeRF++~\cite{zhang2020nerf} takes more than $200$ input images on the Tanks and Temples dataset.

Conventional image-based rendering (IBR)  methods~\cite{choi2019extreme, riegler2020free, hedman2018deep, thies2019image, penner2017soft} first reconstruct a proxy geometry by a multi-view stereo (MVS) algorithm~\cite{huang2018deepmvs, yao2018mvsnet, yao2019recurrent, yang2020cost}.
They then aggregate source views to generate the new view according to the estimated geometry.
Since the two steps are separated from each other, their generated image quality is affected by the accuracy of the reconstructed 3D geometry.

However, developing an end-to-end framework that combines geometry estimation and image synthesis is non-trivial. It requires addressing the following challenges.
First, estimating target view depth by an MVS method will be no longer suitable for end-to-end training because they need to infer depth maps for all source views. It is time- and memory-consuming.
Second, when source view depths are not available, the visibility of target pixels in each source view is hard to infer.
A naive aggregation of warped input images would cause severe image ghosting artifacts.

To tackle the above challenges, we propose to estimate target-view depth and source-view visibility directly from source view images, without estimating depths for source views.
Specifically, we construct a volume under the target view camera frustum.
For each voxel in this volume, when its projected pixel in a source view is similar to the projected pixels in other source views, it is likely that the voxel is visible in this source view.
Motivated by this, we design a source-view visibility estimation module (SVE).
For each source view, our SVE takes the warped source view features as input, compares their similarity with other source views, and outputs visibility of the voxels in this source view.

Then, we aggregate the estimated visibility of the voxels in all source views, obtaining a consensus volume. 
The value in each voxel denotes a surface existence probability. 
% Along each target viewing ray, the surface probability curve might be multi-modal, as shown by the red curve in Fig.~\ref{fig:soft_ray_casting}. 
% In this curve, a smaller peak indicates that a surface is visible by fewer source views, and a larger peak suggests that a surface is visible by a large number of source views. 
Next, we design a soft ray-casting (SRC) mechanism that traverses through the consensus volume along viewing rays and finds the most front surfaces (\ie, depth). 
Since we do not have ground truth target-view depth as supervision, our SRC outputs a depth probability instead of a depth map to model uncertainty. 
% The blue curve in Fig.~\ref{fig:soft_ray_casting} is an example of the output of our SRC algorithm. 

Using the estimated target-view depth and source-view visibility, we warp and aggregate source view pixels to generate the novel view. 
Since the 3D data acquisition is expensive to achieve in practice, 
we do not have any explicit supervision on the depth or visibility. 
Their training signals are provided implicitly by the final image synthesis error. 
We then employ a refinement network to further reduce artifacts and synthesize realistic images. 
To tolerate the visibility estimation error, we feed our refinement network the aggregated images along with warped source view images.

% 
%-------------------------------------------------------------------------
\begin{figure}[t!]
\captionsetup{font={footnotesize}}
\setlength{\abovecaptionskip}{0pt}
    \setlength{\belowcaptionskip}{0pt}
    \centering
    \includegraphics[width=0.8\linewidth]{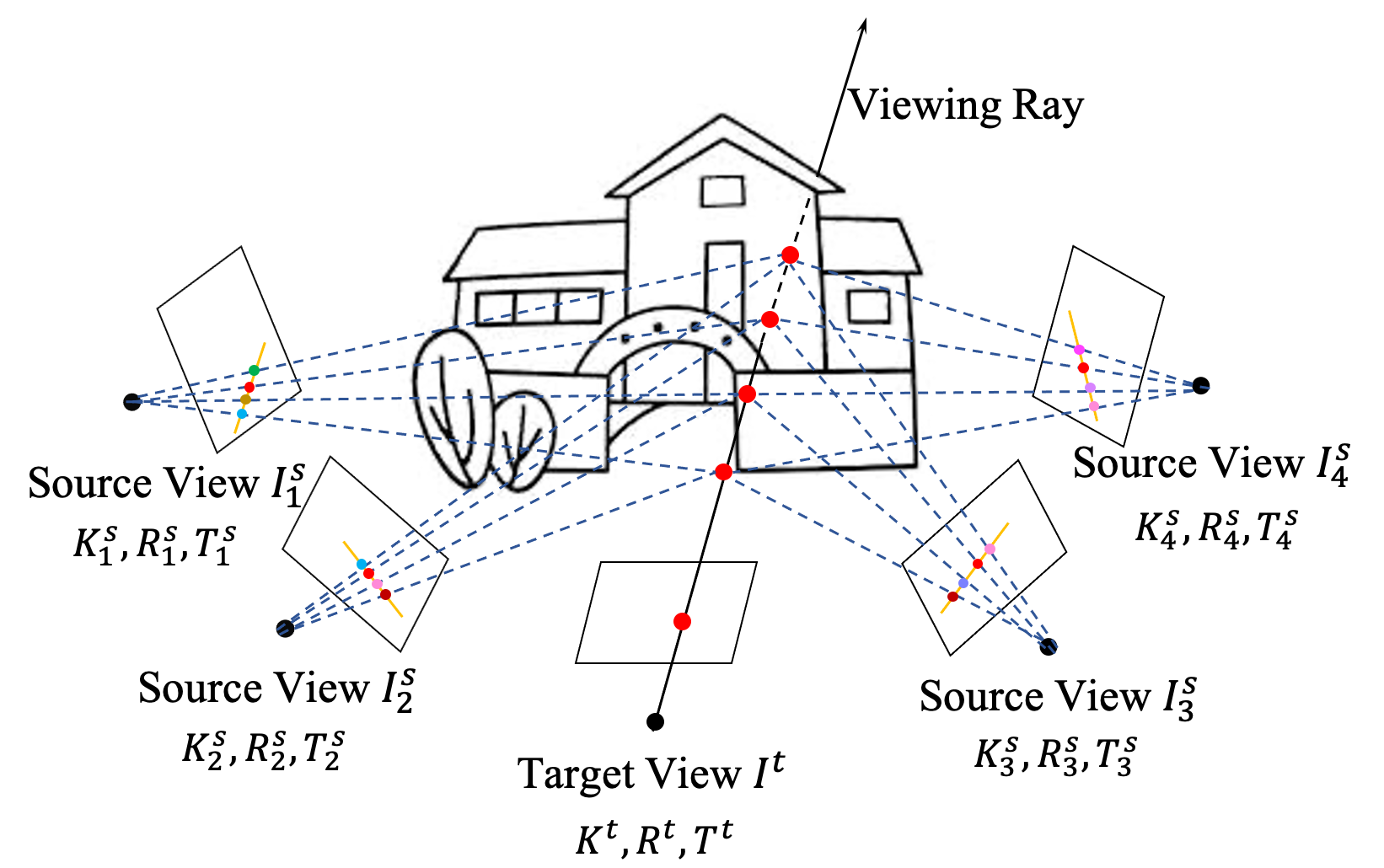}
    \caption{\footnotesize Given a set of unstructured and disordered source views, we aim to synthesize a new view from a position not in the original set. For a 3D point lies in a target viewing ray, when its projected pixels on source view images are consistent with each other, it is of high probability that a surface exists at the corresponding location. The color of this surface can be computed as a visibility-aware combination of source view pixel colors.}
    \label{fig: problem statement}
\end{figure}

\begin{figure*}[!ht!]
\captionsetup{font={footnotesize}}
\setlength{\abovecaptionskip}{0pt}
    \setlength{\belowcaptionskip}{0pt}
    \centering
    \includegraphics[width=0.9\linewidth]{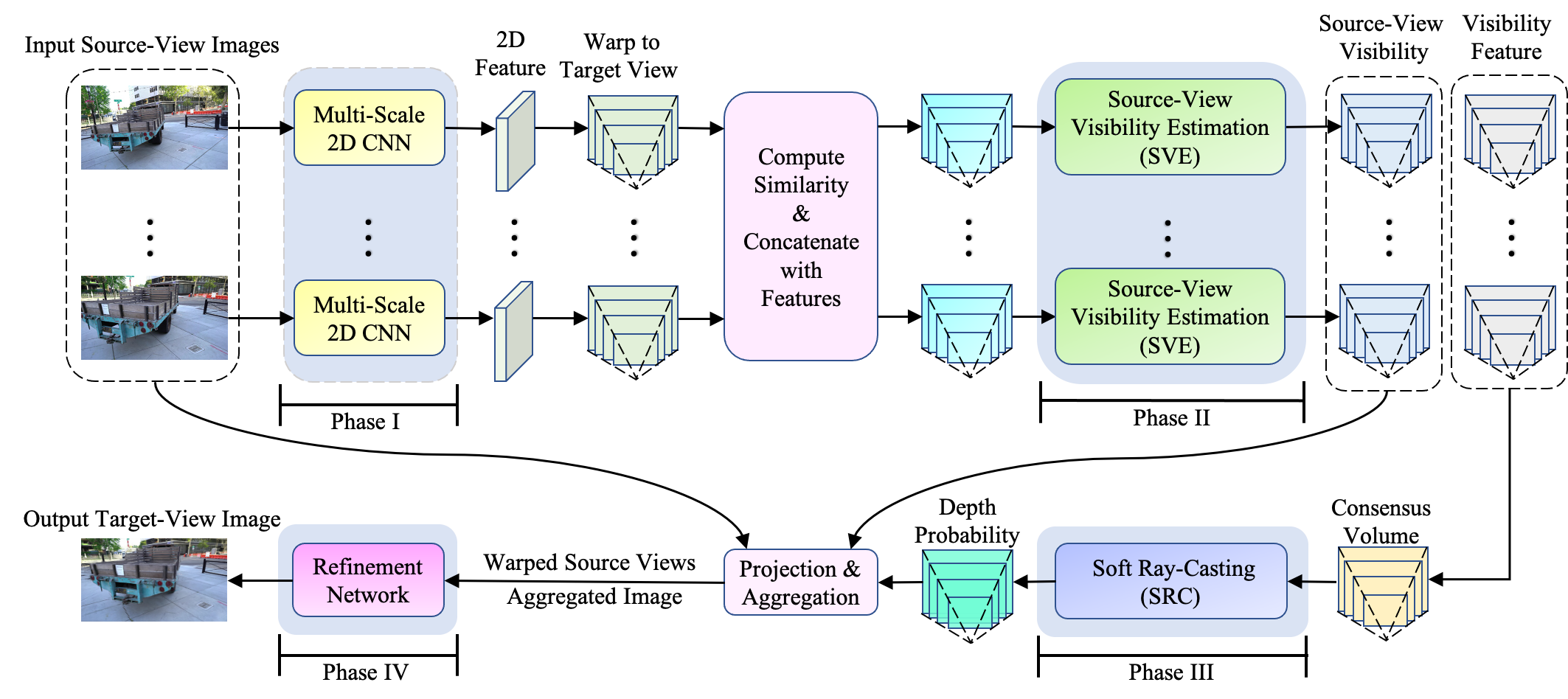}
    \caption{\footnotesize An overall illustration of the proposed framework. We first extract multi-scale features (Phase I) from source images and warp them to the target view. We then design a source-view visibility estimation (SVE) module (Phase II) to estimate the visibility of target voxels in each source view. By aggregating visibility features from all source views, we construct a consensus volume to represent surface existence at different voxels. Next, we design an LSTM-based soft ray-casting (SRC) mechanism (Phase III) to render the depth probability from the consensus volume. By using the estimated source-view visibility and target-view depth, we warp and aggregate source view images. We finally apply a refinement network (Phase IV) to further reduce artifacts in the aggregated images and synthesize realistic novel views. 
    }
    \label{fig:framework}
\end{figure*}
\section{Related Work}

\textbf{Traditional approaches.}
The study of NVS has a long history in the field of computer vision and graphics~\cite{gortler1996lumigraph, levoy1996light, seitz1996view, debevec1996modeling}.
It has important applications in robot navigation, film industry, and augmented/virtual reality~\cite{shi2019spatial, shi2020optimal, shi2020looking, shi2021geometry, hou2020multiview}. 
Buehler \etal~\cite{buehler2001unstructured} define an unstructured Lumigraph and introduce the desirable properties for image-based rendering.
Fitzgibbon \etal~\cite{fitzgibbon2005image} solve the image-based rendering problem as a color reconstruction without explicit 3D geometry modelling.
Penner and Zhang~\cite{penner2017soft} propose a soft 3D reconstruction model that maintains continuity across views and handles depth uncertainty.
% However, the performance of conventional novel view synthesis is often restricted by hand-crafted features.

\textbf{Learning-based methods.}
Recently, learning-based approaches have demonstrated their powerful capability of rendering new views.
Several works have been proposed to train a neural network that learns geometry implicitly and then synthesizes new views~\cite{zhou2016view, sun2018multi, nguyen2020sequential, park2017transformation, eslami2018neural, zhong2018stereo}.
Most of those methods can synthesize arbitrarily new views from limited input views. However, their performance is limited due to the lack of built-in knowledge of scene geometry.

\textbf{Scene representations.}
Some end-to-end novel view synthesis methods model geometry by introducing specific scene representations, such as multi-plane images (MPI)~\cite{zhou2018stereo, mildenhall2019local, srinivasan2019pushing, tucker2020single, flynn2019deepview} and layered depth images (LDI)~\cite{shade1998layered,tulsiani2018layer, shih20203d, tulsiani2018layer}.
MPI represents a scene by a set of front-parallel multi-plane images, and then a novel view image is rendered from it. Similarly, LDI depicts a scene in a layered-depth manner.
% However, the renderable range of MPI and LDI are limited by the depth sampling frequency~\cite{srinivasan2019pushing, shih20203d}.

Deep networks have also been used as implicit functions to represent a specific scene by encapsulating both geometry and appearance from 2D observations~\cite{mildenhall2020nerf, zhang2020nerf, sitzmann2019scene, niemeyer2020differentiable, sitzmann2019deepvoxels, thies2019deferred}.
Those neural scene representations are differentiable and theoretically able to remember all the details of a specific scene. Thus, they can be used to render high-quality images.
However, since these neural representations are used to depict specific scenes, models trained with them are not suitable to synthesize new views from unseen data.

\textbf{Image-based rendering.}
Image-based rendering techniques incorporate geometry knowledge for novel view synthesis. 
They project input images to a target view by an estimated geometry and blend the re-projected images~\cite{choi2019extreme, riegler2020free, hedman2018deep, thies2019image, penner2017soft}. Thus, they can synthesize free-viewpoint images and generalize to unseen data.
However, as geometry estimation and novel view synthesis are two separate steps, these techniques usually produce artifacts when inaccurate geometry or occlusions occur.
Choi \etal~\cite{choi2019extreme} estimate a depth map in the target view by warping source view probability distributions computed by DeepMVS~\cite{huang2018deepmvs}.
To tolerate inaccurate depths, aggregated images as well as original patches extracted from source view images are fed to their proposed refinement network. 
Riegler and Koltun~\cite{riegler2020free} leverage COLMAP~\cite{schonberger2016structure, schonberger2016pixelwise} to reconstruct 3D meshes from a whole sequence of input images and obtain target view depths using the estimated geometry. 
% Although COLMAP is effective for 3D reconstruction, its performance is limited when the number of input views is small.

% Our method can be regarded as an image-based rendering algorithm.
% Different from previous approaches, we implements the geometry estimation and image synthesis in an end-to-end framework. 
% In doing so, the errors from either of these two steps will be eliminated.
% Therefore, our network is able to alleviate the error accumulation issue, thus achieving better view synthesis performance.

\section{Problem Statement}
\label{section:problem statement}

Our goal is to synthesize a novel view $I^t$, given target camera parameters $K^t, R^t, T^t$, from a set of input images, $I_i^s$, $i = 1, 2, ..., N$.
We assume there is sufficient overlap between the source views such that correspondences can be established. We estimate source view camera intrinsic and extrinsic by a well-established structure-from-motion (SfM) pipeline, \eg{COLMAP~\cite{schonberger2016structure}}. 
Fig.~\ref{fig: problem statement} illustrates the situation. 
Mathematically, we formulate this problem as:
% \vspace{-1em}
\begin{equation}
\setlength{\abovedisplayskip}{0pt}
\setlength{\belowdisplayskip}{0pt}
\small
\begin{split}
    I^{t*} &= \argmax_{I^t}~ p\left (I^t | I_1^s, I_2^s, ..., I_N^s \right ),
        %   &= \argmax_{I^t}~~ \prod_{(u, v)}p\left (I^t (u, v)| I_1^s, I_2^s, ..., I_N^s  \right ),
\end{split}
\label{eq: problem statement}
\end{equation}
% \vspace{-5em}
where $p(\cdot)$ is a probability function.

Due to the expensive accessibility of 3D data (\eg, depths) and a limited number of input views, it is hard to compute accurate 3D geometry from input source views. 
Therefore, our intuition is to develop an end-to-end framework that combines geometry estimation and image synthesis, to eliminate the error propagation issue.
We achieve this goal by estimating target-view depth and source-view visibility for target pixels directly under the target view.  

We assume a uniform prior on the target view depth, and reformulate Eq.~\eqref{eq: problem statement} as a probability conditioned on depth $d$:
\begin{equation}
\setlength{\abovedisplayskip}{0pt}
\setlength{\belowdisplayskip}{0pt}
\small
\begin{split}
    I^{t*} &= \argmax_{I^t}~\sum_{d=d_\text{min}}^{d_\text{max}}p\left (I^t |d \right )p\left ( d \right )\\
           &= \sum_{d=d_\text{min}}^{d_\text{max}} \left [ \argmax_{I^t}~p\left (I^t |d \right ) \right] p(d),
\end{split}
\label{eq:condition_on_depth}
\end{equation}
where $d_\text{min}$ and $d_\text{max}$ are statistical minimum and maximum depths of a target view respectively. 
As the source view images are given, we omit them in this equation.

Following conventional methods, we compute the target view color $I^t$ with the highest probability given depth $d$ as a visibility-aware combination of source view colors:
\begin{equation}
\setlength{\abovedisplayskip}{0pt}
\setlength{\belowdisplayskip}{0pt}
\small
\begin{split}
    \argmax_{I^t}~p\left (I^t|d \right )
    % = I^{t} (u, v|d) 
    %  = \argmax_{I^t}~p\left (I^t |d \right )
     = \sum_{i=1}^{N}\mathbf{w}_i^d \mathbf{C}_i^d,
    %  ~~ s.t.\sum_{i=1}^{N}w_i^d=1,
    \label{eq:aggregation}
\end{split}
\end{equation}
where $\mathbf{C}_i^d \in \mathbb{R}^{H \times W \times 3}$ is a collection of re-projected target pixels in source view $i$ by inverse warping~\cite{li2006inverse},  
$\mathbf{w}_i^d\in \mathbb{R}^{H \times W}$ is the blending weight of source view $i$, 
and it is computed from the visibility of target pixels in each source view:
\begin{equation}
\setlength{\abovedisplayskip}{0pt}
\setlength{\belowdisplayskip}{0pt}
\small
    \mathbf{w}_i^d = exp{\left ( \mathbf{V}_i^d\right )}/\sum_{i=1}^{N}exp \left (\mathbf{V}_i^d \right) ,
\end{equation}
where $\mathbf{V}_i^d \in \mathbb{R}^{H\times W}$ is the visibility of target pixels in source view $i$ given the target-view depth $d$. 

In the next section, we will provide technical details on how to estimate the source-view visibility $\mathbf{V}_i^d$ and target-view depth probability distribution $p(d)$.

\section{The Proposed Framework}

We aim to construct an end-to-end framework for novel view synthesis from a few sparse input images. 
By doing so, inaccurately-estimated geometry can be corrected by image synthesis error during training. 
We achieve this goal by estimating target-view depth and source-view visibility directly under the target view. 
Fig.~\ref{fig:framework} depicts the proposed pipeline.

Start from a blank volume in the target-view camera frustum. 
Our goal is to select pixels from source-view images to fill in the voxels of this volume. 
After that, we can render the target view image from this colored volume. 
In this process, the visibility of the voxels in each source view, and the target-view depth, are two of the most crucial issues.

\subsection{A multi-scale 2D CNN to extract features}

When a voxel of this volume is visible in a source view, its projected pixel in this source view should be similar to the projected pixels in other source views. 
This is the underlying idea for the source-view visibility estimation.
However, the pixel-wise similarity measure is not suitable for textureless and reflective regions. 
Hence, we propose to extract high-level features from source view images for the visibility estimation by a 2D CNN.

Our 2D CNN includes a set of dilation convolutional layers with dilation rates as 1, 2, 3, 4 respectively. 
Its output is a concatenation of extracted multi-scale features. 
This design is to increase the receptive field of extracted features and retain low-level detailed information~\cite{yan2020dense}, thus increasing discriminativeness for source view pixels.

Denote $\mathbf{F}_i \in \mathbb{R}^{H \times W \times D \times C} $ as 
the warped features of source view $i$, where $D$ is the sampled depth plane number in the target view. 
For target-view voxels at depth $d$, we compute the similarity between corresponding features in source view $i$ and other source views as:
\begin{equation}
\setlength{\abovedisplayskip}{0pt}
\setlength{\belowdisplayskip}{0pt}
\small
    \mathbf{S}_i^d = \sum_{j=1, j\neq i}^{N} {Sim}\left (  \mathbf{F}_i^d, \mathbf{F}_j^d \right )/(N-1),
    \label{eq:sim}
\end{equation}
where $Sim(\cdot, \cdot)$ is a similarity measure between its two inputs, and we adopt cross-correlation in this paper.

\subsection{Source-view visibility estimation (SVE) module}
In theory, deep networks are able to learn viewpoint invariant features. 
However, we do not have any explicit supervision on the warped source-view features. 
The computation of source-view visibility for the voxels is too complex to be modeled by Eq.~\eqref{eq:sim}. 
Hence, we propose to learn a source-view visibility estimation (SVE) module to predict the visibility of the voxels in each source view. 

Our SVE module is designed as an encoder-decoder architecture with an LSTM layer at each stage. The LSTM layer is adopted to encode sequential information along the depth dimension. 
Our SVE module takes into account self-information ($ \mathbf{F}_i^d$), local information ($ \mathbf{S}_i^d$) and global information ($\sum_{i=1}^{N}\mathbf{S}_i^d/N$) to determine the visibility of the voxels in each source view. 
Mathematically, we express it as:
\begin{equation}
\setlength{\abovedisplayskip}{0pt}
\setlength{\belowdisplayskip}{0pt}
\small
    {\mathbf{V}}_i^d, \mathbf{B}_i^d, \text{state}_f^d = f\left ( \left [ \mathbf{F}_i^d, \mathbf{S}_i^d, 
    \sum_{i=1}^{N}\mathbf{S}_i^d/N
    \right ],  \text{state}_f^{d-1}\right),
\end{equation}
where $[\cdot]$ is a concatenation operation, $\mathbf{V}_{i}^{d}\in \mathbb{R}^{H\times W}$ is the estimated visibility for target-view voxels at depth $d$ in source view $i$, $ \mathbf{B}_i^d \in \mathbb{R}^{H\times W \times 8}$ is the associated visibility feature, $f(\cdot)$ denotes the proposed SVE module, $\text{state}_{f}^{d-1}$ is the past memory of our SVE module before depth $d$ and $\text{state}_{f}^{d}$ is the updated memory at depth $d$.

\begin{figure}[t!]
\captionsetup{font={footnotesize}}
\setlength{\abovecaptionskip}{0pt}
    \setlength{\belowcaptionskip}{0pt}
    \centering
    \includegraphics[width=\linewidth]{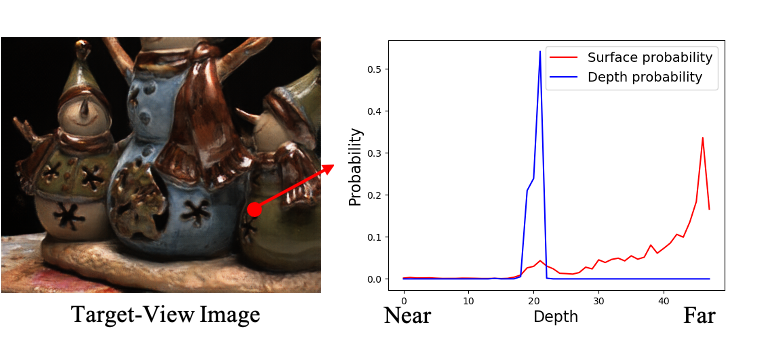}
    \caption{\footnotesize Illustration of our soft ray-casting mechanism. 
    Our SRC traverses through the surface probability curve along target viewing rays from near to far, increases the depth probability of the first opaque element, and decreases depth probabilities of later elements no matter they are opaque or not.
    }
    \label{fig:soft_ray_casting}
\end{figure}

\subsection{Soft ray-casting (SRC) mechanism} 
By aggregating visibility from all source views, we obtain a surface existence probability for each voxel in the target view.
As shown in Fig.~\ref{fig:soft_ray_casting}, the surface probability curve along a target viewing ray might be multi-modal, where a smaller peak indicates that a surface is visible by fewer source views and a larger peak suggests that its corresponding surface is visible by a large number of source views. 

To increase the representative ability, we aggregate the visibility features, instead of visibility, of source views to compute a consensus volume:
\begin{equation}
\setlength{\abovedisplayskip}{0pt}
\setlength{\belowdisplayskip}{0pt}
\small
    \mathcal{C} =  \sum_{i=1}^{N}\mathbf{B}_i/N,
    \label{eq:consensus volume}
\end{equation}
where $\mathcal{C}\in \mathbb{R}^{H \times W \times D \times 8}$ is the obtained consensus volume. 

Then, we design a soft ray-casting (SRC) mechanism to render the target view depth from the consensus volume. 
Our SRC is implemented in the form of an LSTM layer. 
Similar to our SVE module, the LSTM layer is to encode sequential relationship along the depth dimension.

The LSTM layer traverses though the consensus volume along target viewing rays from near to far. 
When meeting the most front surface, it outputs a large depth probability value for the corresponding voxel. 
For the later voxels, the LSTM layer sets their probability values to zero. 
Denote the LSTM cell as $r(\cdot)$.
At each depth $d$, it takes as input the current consensus feature $\mathcal{C}^d$ and its past memory $\text{state}_r^{d-1}$, and outputs the depth probability $p(d)$ along with the updated memory $\text{state}_r^{d}$:
\begin{equation}
\setlength{\abovedisplayskip}{0pt}
\setlength{\belowdisplayskip}{0pt}
\small
    \text{state}_r^{d}, p(d) = r(\mathcal{C}^d, \text{state}_r^{d-1}).
\end{equation}

\begin{figure*}
\captionsetup{font={footnotesize}}
\setlength{\abovecaptionskip}{0pt}
    \setlength{\belowcaptionskip}{0pt}
    \centering
    \subfigure[EVS~\cite{choi2019extreme}]{
    \begin{minipage}{0.235\linewidth}
    \includegraphics[width=\linewidth]{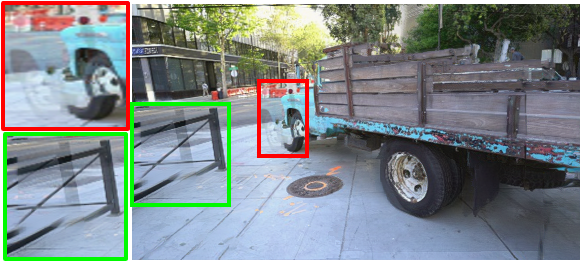}\\
    \includegraphics[width=\linewidth]{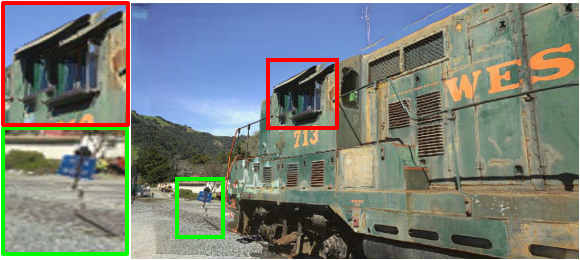}\\
    \includegraphics[width=\linewidth]{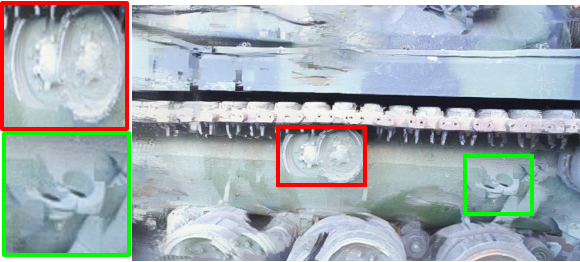}\\
    \includegraphics[width=\linewidth]{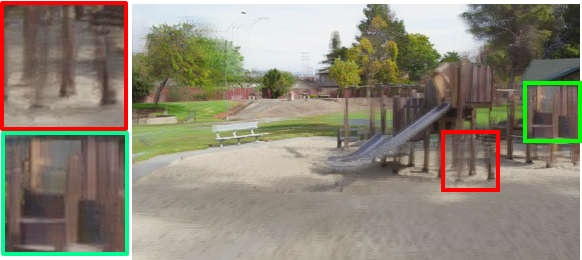}
    \end{minipage}
    \label{subfig:EVS}
    }
    \subfigure[FVS~\cite{riegler2020free}]{
    \begin{minipage}{0.235\linewidth}
    \includegraphics[width=\linewidth]{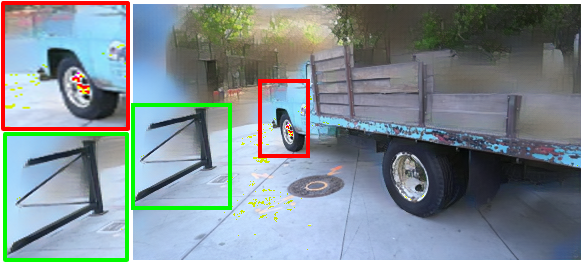}\\
    \includegraphics[width=\linewidth]{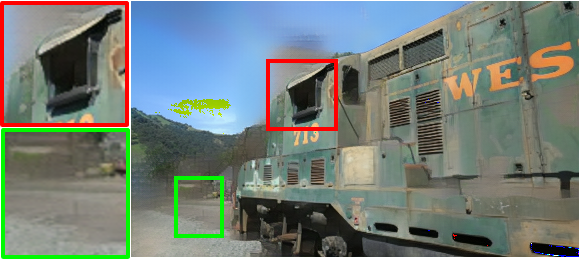}\\
    \includegraphics[width=\linewidth]{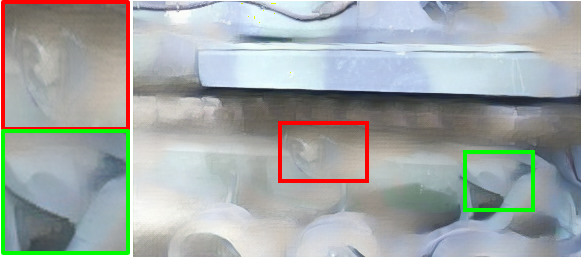}\\
    \includegraphics[width=\linewidth]{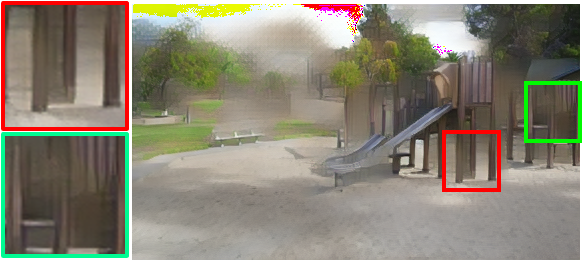}
    \end{minipage}
    \label{subfig:FVS}
    }
    \subfigure[Ours]{
    \begin{minipage}{0.235\linewidth}
    \includegraphics[width=\linewidth]{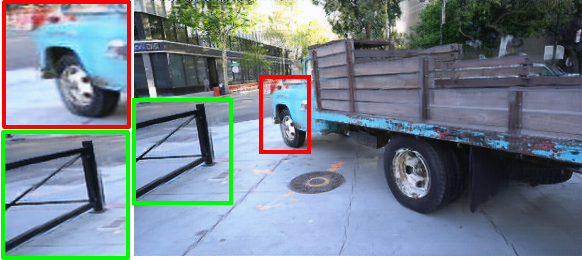}\\
    \includegraphics[width=\linewidth]{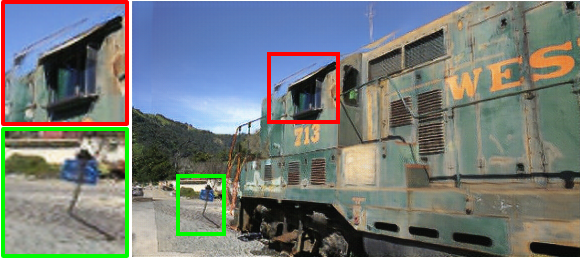}\\
    \includegraphics[width=\linewidth]{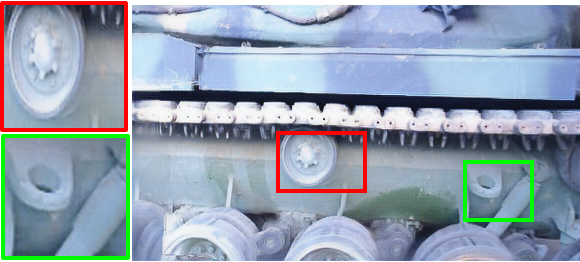}\\
    \includegraphics[width=\linewidth]{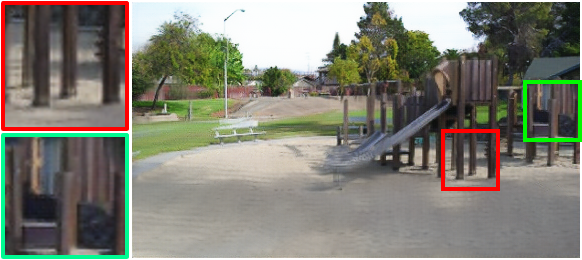}
    \end{minipage}
    \label{subfig:ours}
    }
    \subfigure[Ground Truth]{
    \begin{minipage}{0.235\linewidth}
    \includegraphics[width=\linewidth]{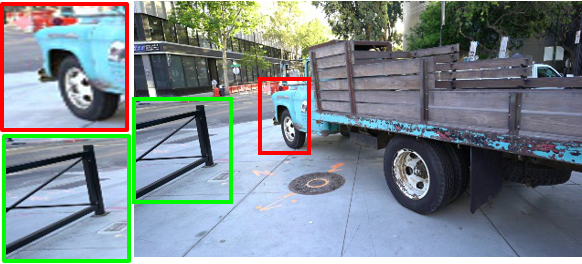}\\
    \includegraphics[width=\linewidth]{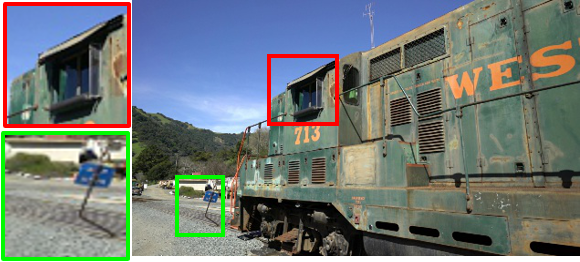}\\
    \includegraphics[width=\linewidth]{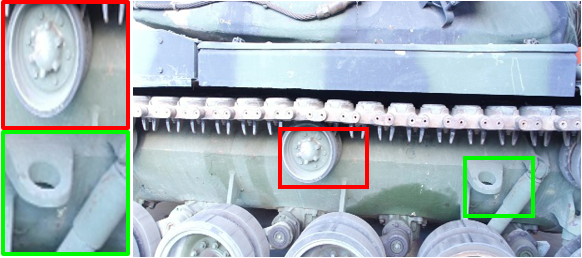}\\
    \includegraphics[width=\linewidth]{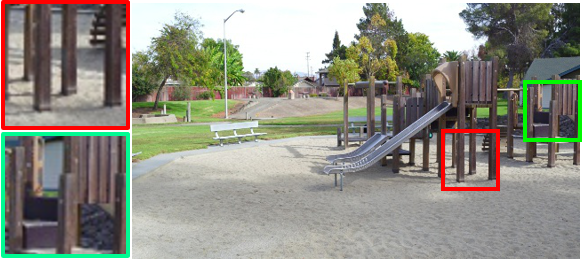}
    \end{minipage}
    \label{subfig:GT}
    }
    \caption{\footnotesize Qualitative visualization of generated results on the Tanks and Temples dataset with six views as input. The four examples are from scene \textit{Truck}, \textit{Train}, \textit{M60}, \textit{Playground} respectively. }
    \label{fig:stoa_tat}
\end{figure*}

\begin{table*}[ht!]
\captionsetup{font={footnotesize}}
\setlength{\abovecaptionskip}{0pt}
    \setlength{\belowcaptionskip}{0pt}
\setlength{\tabcolsep}{2pt}
\centering
\caption{\footnotesize Quantitative comparison with the state-of-the-art. Here, ``Whole'' denotes using the whole sequence as input; ``*'' indicates that results are from Zhang~\etal~\cite{zhang2020nerf}; and ``$^\dagger$ '' represents that results are from Riegler and Koltun~\cite{riegler2020free}. }
\footnotesize
\begin{tabularx}{\linewidth}{c|c|ccc|ccc|ccc|ccc|ccc}
\toprule
& \multirow{3}{*}{\begin{tabular}[c]{@{}c@{}}Input\\ View \\Number \end{tabular}} & \multicolumn{12}{c|}{Tanks and Temples}                                                                                                                                                                    & \multicolumn{3}{c}{\multirow{2}{*}{DTU}}         \\
                                     &                                         & \multicolumn{3}{c|}{Truck}                        & \multicolumn{3}{c|}{Train}                        & \multicolumn{3}{c|}{M60}                          & \multicolumn{3}{c|}{Playground}                   & \multicolumn{3}{c}{}                             \\ 
                                     &                                         & LPIPS$\downarrow$         & SSIM$\uparrow$           & PSNR$\uparrow$           & LPIPS$\downarrow$          & SSIM$\uparrow$           & PSNR$\uparrow$           & LPIPS$\downarrow$          & SSIM$\uparrow$           & PSNR$\uparrow$           & LPIPS$\downarrow$          & SSIM$\uparrow$           & PSNR$\uparrow$           & LPIPS$\downarrow$          & SSIM$\uparrow$           & PSNR$\uparrow$           \\ \midrule
EVS\cite{choi2019extreme}  & 6 & 0.301          & 0.588          & 17.74          & 0.434          & 0.434          & 15.38          & 0.314          & 0.585          & 16.40          & \textbf{0.243} & 0.702          & 21.57          & 0.32          & 0.645          & 17.83          \\
FVS\cite{riegler2020free} & 6 & 0.318          & 0.638          & 15.82          & 0.447          & 0.502          & 13.71          & 0.486          & 0.548          & 11.49          & 0.417          & 0.586          & 14.95          & 0.47          & 0.530          & 10.45          \\ 
Ours&  6 & \textbf{0.233} & \textbf{0.708} & \textbf{21.33} & \textbf{0.386} & \textbf{0.542} & \textbf{18.81} & \textbf{0.250} & \textbf{0.732} & \textbf{19.20} & 0.245          & \textbf{0.710} & \textbf{22.12} & \textbf{0.303} & \textbf{0.721} & \textbf{19.20}   \\ 
\midrule
\multirow{3}{*}{Ours} 
 
 &2                                                                              & 0.294          & 0.627          & 19.20          & 0.475          & 0.464          & 17.44          & 0.303          & 0.667          & 18.14          & 0.350          & 0.604          & 19.98          & 0.362          & 0.625          & 17.54          \\
  & 4                                                                              & 0.254          & 0.682          & 20.41          & 0.430          & 0.495          & 17.82          & 0.283          & 0.698          & 19.37          & 0.275          & 0.663          & 21.31          & 0.359          & 0.646          & 17.84          \\
  &  6 & \textbf{0.233} & \textbf{0.708} & \textbf{21.33} & \textbf{0.386} & \textbf{0.542} & \textbf{18.81} & \textbf{0.250} & \textbf{0.732} & \textbf{19.20} & \textbf{0.245}          & \textbf{0.710} & \textbf{22.12} & \textbf{0.303} & \textbf{0.721} & \textbf{19.20}   \\ 
  \midrule  %\cmidrule{2-17}

NeRF\cite{mildenhall2020nerf}  & Whole & 0.513*          & 0.747*          & 20.85*          & 0.651*          & 0.635*          & 16.64*          & 0.602*          & 0.702*          & 16.86*          & 0.529* & 0.765*          & 21.55*          & --          & --          & --         \\
FVS\cite{riegler2020free} & Whole & 0.11$^\dagger$          & 0.867$^\dagger$           & 22.62$^\dagger$           & 0.22$^\dagger$           & 0.758$^\dagger$           & 17.90$^\dagger$           & 0.29$^\dagger$           & 0.785$^\dagger$           & 17.14$^\dagger$           & 0.16$^\dagger$           & 0.837$^\dagger$           & 22.03$^\dagger$           & --          & --          & --          \\ 
 \bottomrule
\end{tabularx}
\label{tab:stoa}
\end{table*}

\subsection{Refinement network}
Using the estimated source-view visibility and target-view depth probability, we aggregate the source images and obtain $I^{t*}$ by Eq.~\eqref{eq:condition_on_depth}. 
We then employ a refinement network to further reduce artifacts on the aggregated image.

Our refinement network is designed in an encoder-decoder architecture with convolutional layers. 
To tolerate errors caused by the visibility estimation block, the encoder in our refinement network is in two branches: one for the aggregated image $I^{t*}$ and another for a warped source view $I_i^{\text{warp}} = \sum_{d=1}^D \mathbf{C}_i^d p(d)$. 
Its outputs are a synthesized target view image $\hat{I^t_i}$ along with a confidence map $m_i$:
\begin{equation}
\setlength{\abovedisplayskip}{0pt}
\setlength{\belowdisplayskip}{0pt}
\small
    \hat{I^t_i}, m_i = \text{Refinement}(I^{t*}, I_i^{\text{warp}}).
\end{equation}
The final output of our refinement network is computed as:
\begin{equation}
\setlength{\abovedisplayskip}{0pt}
\setlength{\belowdisplayskip}{0pt}
\small
    \widetilde{I^{t}} = \sum_{i=1}^{N}m_i \hat{I_i^t}.
\end{equation}

\subsection{Training objective}
We employ the GAN training scheme to train our framework. 
For brevity, we omit the adversarial loss in this paper. Interested readers are referred to Isola~\etal~\cite{isola2017image}. 
For the target image supervision, we adopt the perceptual loss of Chen and Koltun~\cite{chen2017photographic}:
\begin{equation}
\setlength{\abovedisplayskip}{0pt}
\setlength{\belowdisplayskip}{0pt}
\small
    \mathcal{L}_{per} = \left \| \widetilde{I^{t}} - I^t  \right \|_1 + 
\sum_l \lambda_l\left \| \phi_l(\widetilde{I^{t}}) - \phi_l(I^t) \right \|_1,
\end{equation}
where $\phi(\cdot)$ indicates the outputs of a set of layers from a pretrained VGG-19~\cite{simonyan2014very}, and $\left \| \cdot \right \|_1$ is the $L_1$ distance. 
The settings for coefficients $\lambda_l$ are the same as Zhou~\etal~\cite{zhou2018stereo}.

\textbf{Self-supervised training signal for our SRC and SVE. }
Generally, it is difficult for our SRC to decide which is the most front surface in a viewing ray, especially when the surface probability curve is multi-modal. 
We expect this soft determination can be learned statistically from training.
Particularly, when the estimated depth is incorrect, the color of warped pixels from source-view images will deviate from the ground truth target view color. 
This signal would punish the LSTM and helps it to make the right decisions. 
The same self-supervised training scheme is applied to our SVE module. 
We illustrate the estimated depth for a target pixel in Fig.~\ref{fig:soft_ray_casting}, and an example of the visibility-aware aggregated image in Fig.~\ref{fig:visibility}.

\section{Experiments}

\begin{figure*}
\captionsetup{font={footnotesize}}
\setlength{\abovecaptionskip}{0pt}
    \setlength{\belowcaptionskip}{0pt}
    \centering
    \subfigure[EVS~\cite{choi2019extreme}]{
    \begin{minipage}{0.235\linewidth}
    \includegraphics[width=\linewidth]{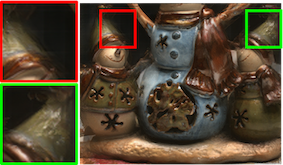}\\
    \includegraphics[width=\linewidth]{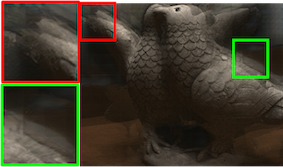}
    \end{minipage}
    }
    \subfigure[FVS~\cite{riegler2020free}]{
    \begin{minipage}{0.235\linewidth}
    \includegraphics[width=\linewidth]{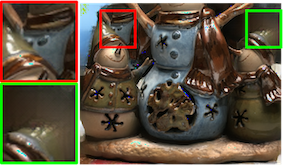}\\
    \includegraphics[width=\linewidth]{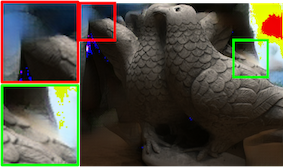}
    \end{minipage}
    }
    \subfigure[Ours]{
    \begin{minipage}{0.235\linewidth}
    \includegraphics[width=\linewidth]{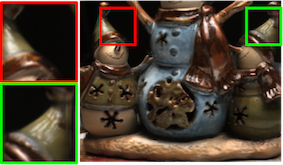}\\
    \includegraphics[width=\linewidth]{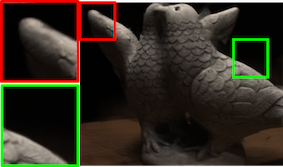}
    \end{minipage}
    }
    \subfigure[Ground Truth]{
    \begin{minipage}{0.235\linewidth}
    \includegraphics[width=\linewidth]{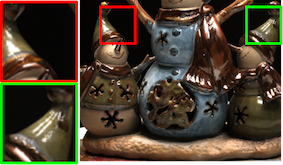}\\
    \includegraphics[width=\linewidth]{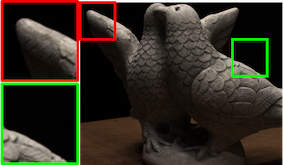}
    \end{minipage}
    }
    \caption{\footnotesize Qualitative visualization of generated results on the DTU dataset with six views as input.
    The two examples are from scene \textit{scan3} and \textit{scan106}, respectively.}
    \label{fig:stoa_dtu}
\end{figure*}

\textbf{Dataset and evaluation metric.}
We conduct experiments on two datasets, {Tanks and Temples}~\cite{knapitsch2017tanks}, and {DTU}~\cite{aanaes2016large}. Camera movements in the two datasets include both rotations and translations.

On the {Tanks and Temples}, we use the training and testing split provided by Riegler and Koltun~\cite{riegler2020free}. In this dataset, $17$ out of $21$ scenes are selected out as the training set. The remaining four scenes, \textit{Truck}, \textit{Train}, \textit{M60}, and \textit{Playground}, are employed as the testing set. 
We apply the leave-one-out strategy for training, namely, designating one of the images as target image and selecting its nearby $N$ images as input source images. 
For testing, different from Riegler and Koltun~\cite{riegler2020free} which uses whole sequences as input, we select $N$ nearby input images for each target view.

For the {DTU} dataset, it is employed to further demonstrate the generalization ability of trained models. We do not train on this dataset and use the validation set provided by Yao~\etal~\cite{yao2018mvsnet}. 
The validation set includes 18 scenes. Each of the scenes contains 49 images. We apply the same leave-one-out strategy as on the Tanks and Temples dataset.

Following recent NVS works~\cite{choi2019extreme, riegler2020free, liu2018geometry, flynn2019deepview}, we adopt the commonly used SSIM, PSNR and LPIPS~\cite{zhang2018unreasonable} for quality evaluation on synthesized images. 

\textbf{Implementation details.}
% We adopt an Adam optimizer with learning rate as $2\times 10^{-4}$ and set $\beta_1 = 0.5$, $\beta_2 = 0.999$, $\epsilon=10^{-8}$. 
For the Tanks and Temples, we experiment on image resolution of $256\times 448$. 
For the DTU dataset, the input image resolution is $256\times 320$. 
We use a TITAN V with 12GB memory to train and evaluate our models. We train $10$ epochs on the Tanks and Temples dataset with a batch size of $1$. It takes $20$ hours for training using $6$ input images, and $0.35s$ per image (average) for evaluation.
We apply inverse depth sampling strategy with depth plane number $D=48$. For outdoor scenes, \ie, the Tanks and Temples, we set $d_\text{min}=0.5$m and $d_\text{max}=100$m. 
For constrained scenes, \ie, the DTU dataset, we employ the minimum ($425$mm) and maximum ($937$mm) depth in the whole dataset. 
The source code of this paper is available at \url{https://github.com/shiyujiao/SVNVS.git}.
% Please refer to our supplementary material for the network architectures. 
% The source code with all network architecture details will be released to facilitate reproducible research.

\subsection{Comparison with the state-of-the-art}
We first compare with two recent and representative IBR methods, Extreme View Synthesis (EVS)~\cite{choi2019extreme} and Free View Synthesis (FVS)~\cite{riegler2020free}, with six views as input.
We present the quantitative evaluation in the first three rows of Tab.~\ref{tab:stoa}. Qualitative comparisons on the Tanks and Temples are presented in Fig.~\ref{fig:stoa_tat}.

Both EVS and FVS first estimate depth maps for source views.
In their methods, the visibility of target pixels in source views is computed by a photo-consistency check between source and target view depths.
EVS aggregates source views simply based on the source-target camera distance.
Their aggregation weights do not have the ability to tolerate visibility error caused by in-accurate depth.
Thus, the synthesized images by EVS suffer severe ghosting artifacts, as shown in Fig.~\ref{subfig:EVS}.
FVS employs a COLMAP to reconstruct the 3D mesh.
When input images densely cover a scene, the reconstructed geometry is exceptionally good, and the synthesized images are of high-quality, as shown in the last row of Tab.~\ref{tab:stoa}.
However, when the number of input images are reduced, \ie, $6$, the reconstructed mesh by COLMAP is of poor quality, and the depth-incorrect regions in the synthesized images are blurred, as indicated in~\ref{subfig:FVS}.
In contrast, our method does not rely on the accuracy of estimated source-view depths or reconstructed 3D mesh.
Instead, we directly recover target-view depth and source-view visibility from input images.
Thus, our synthesized images show higher quality than the recent state-of-the-art.

\textbf{Generalization ability.} To further demonstrate the generalization ability, we employ the trained models of the three algorithms to test on the DTU dataset. Quantitative results are presented in the last column of Tab.~\ref{tab:stoa}. Our method consistently outperforms the recent state-of-the-art algorithms. We present two visualization examples in Fig.~\ref{fig:stoa_dtu}. More qualitative results are provided in the supplementary material.

\textbf{Different input view number.}
We further conduct experiments on reducing the number of input views of our method. Quantitative results are presented in the bottom part of Tab.~\ref{tab:stoa}. Increasing the input view number improves the quality of synthesized images. This conforms to our general intuition that image correspondences can be easily established and more disoccluded regions can be observed when more input views are available.

\textbf{Comparison with NeRF.}
For completeness, we present the performance of NeRF~\cite{mildenhall2020nerf} with the whole sequence as input in the penultimate row of Tab.~\ref{tab:stoa}.
The major difference between NeRF and our method is the different problem settings.
NeRF is more suitable to view synthesis on a specific scene with many images as input.
When the scene is changed, NeRF needs to be re-trained on the new scene.
In contrast, we expect our network to learn common knowledge from its past observations (training data) and be able to apply the learned knowledge to unseen scenes without further fine-tuning.
Thus, our approach is a better choice when the trained model is expected to generalize, and the number of input images is small.  

\textbf{Comparison with Szeliski and Golland~\cite{szeliski1999stereo}.}
We found our work shares the same spirit with a classical work~\cite{szeliski1999stereo}.
Both works construct a virtual camera frustum under the target view and aim to estimate the color and density (depth probability in our work) for each of its elements.
Szeliski and Golland~\cite{szeliski1999stereo} first compute an initial estimation by finding agreement among source views.
Next, they project the estimation to the source views, compute the visibility of source views, and then refine the estimation iteratively.
Benefited from learning-based techniques, our approach encodes the visibility estimation as a single forward step (compared to iterative refinement in Szeliski and Golland~\cite{szeliski1999stereo}) and can handle more complex scenarios, such as textures and reflective regions, as shown in the qualitative visualizations of this paper and supplementary material.

\begin{figure*}
\captionsetup{font={footnotesize}}
\setlength{\abovecaptionskip}{0pt}
    \setlength{\belowcaptionskip}{0pt}
    \centering
    \begin{minipage}{0.69\linewidth}
    \subfigure[Warped Source-View Images in Target View]{
    \begin{minipage}{\linewidth}
    \vspace{0.25em}
    \includegraphics[width=0.32\linewidth]{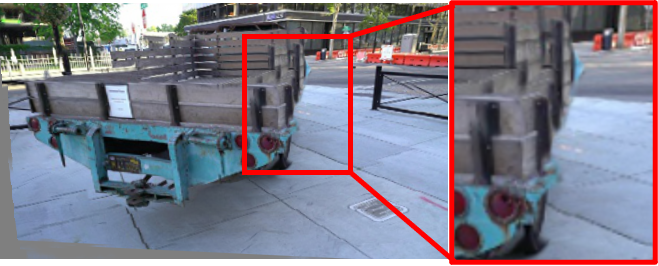}
    \includegraphics[width=0.32\linewidth]{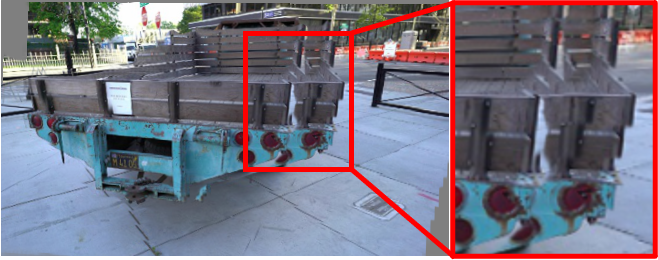}
    \includegraphics[width=0.32\linewidth]{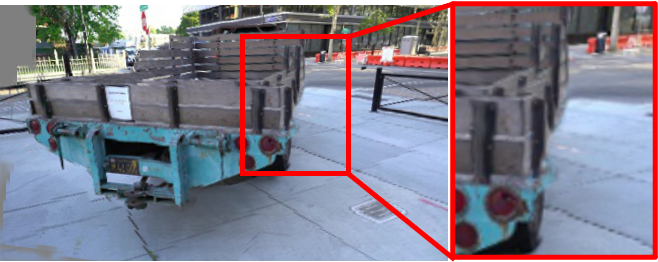}
    \vspace{1.5em}\\
    \includegraphics[width=0.32\linewidth]{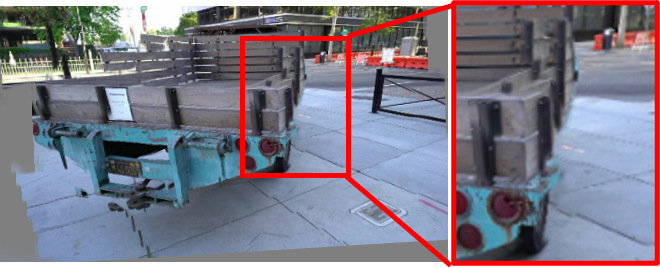}
    \includegraphics[width=0.32\linewidth]{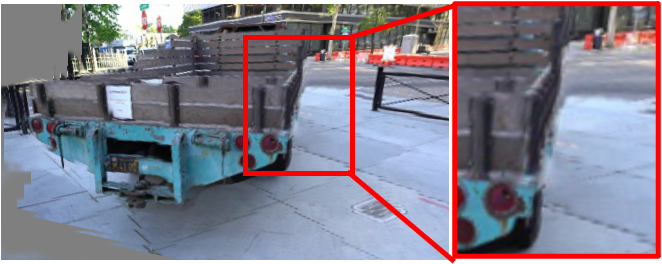}
    \includegraphics[width=0.32\linewidth]{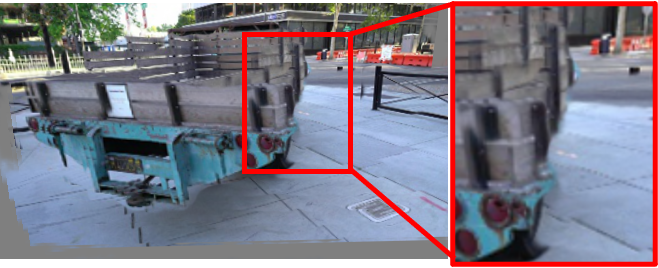}\\
    \vspace{-1em}
    \end{minipage}
    \label{subfig:visibility_warped_source_view}
    }
    \end{minipage}
    \begin{minipage}{0.2208\linewidth}
    \vspace{-0.5em}
    \subfigure[Aggregated Image]{
    \includegraphics[width=\linewidth]{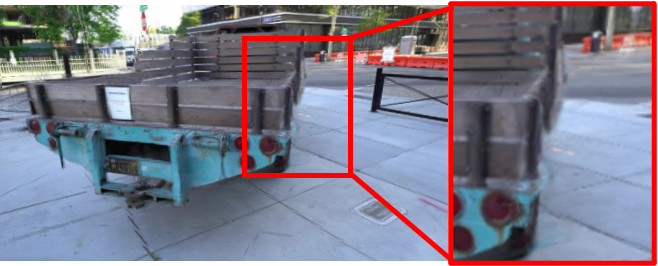}
    \label{subfig:visibility_aggregated_source_view}
    }\\
    \vspace{-1em}
    \subfigure[Ground Truth]{
    \includegraphics[width=\linewidth]{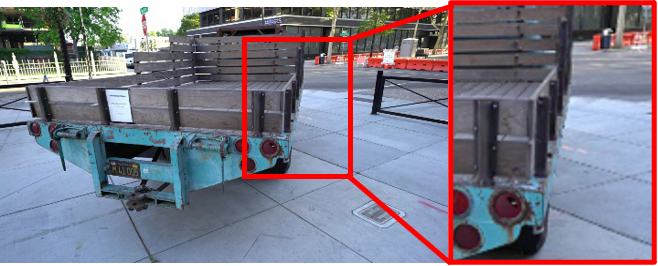}
    \label{subfig:visibility_gt}
    \vspace{-10em}
    }\\
    \vspace{-1em}
    \end{minipage}
    \caption{ \small Qualitative illustration on our visibility-aware aggregation. (a) Warped source images in target view. There are severe ghosting artifacts due to occlusions. (b) Aggregated image by using our visibility-aware blending weights. (c) Target view ground truth image.}
    \label{fig:visibility}
\end{figure*}

\begin{table*}
\captionsetup{font={footnotesize}}
\setlength{\abovecaptionskip}{0pt}
    \setlength{\belowcaptionskip}{0pt}
\setlength{\tabcolsep}{1.5pt}
\centering
\caption{\footnotesize Necessity of each module in the proposed framework.}
\footnotesize
\begin{tabularx}{\linewidth}{l|ccc|ccc|ccc|ccc|ccc}
\toprule
\multirow{3}{*}{}       & \multicolumn{12}{c|}{Tanks and Temples}                                                                                                                                                                    & \multicolumn{3}{c}{\multirow{2}{*}{DTU}}         \\
                        & \multicolumn{3}{c|}{Truck}                        & \multicolumn{3}{c|}{Train}                        & \multicolumn{3}{c|}{M60}                          & \multicolumn{3}{c|}{Playground}                   & \multicolumn{3}{c}{}                             \\
                        & LPIPS$\downarrow$          & SSIM$\uparrow$          & PSNR$\uparrow$           & LPIPS$\downarrow$          & SSIM$\uparrow$           & PSNR$\uparrow$           & LPIPS$\downarrow$          & SSIM$\uparrow$           & PSNR$\uparrow$           & LPIPS$\downarrow$          & SSIM$\uparrow$           & PSNR$\uparrow$           & LPIPS$\downarrow$          & SSIM$\uparrow$           & PSNR$\uparrow$           \\ \midrule
Ours w/o visibility     & 0.271          & 0.660          & 20.35          & 0.437          & 0.482          & 17.93          & 0.322          & 0.656          & 16.83          & 0.277          & 0.665          & 21.40          & 0.367          & 0.641          & 17.03          \\ 
Ours w/o ray-casting    & 0.761          & 0.643          & 20.04          & 0.665          & 0.487          & 17.83          & 0.766          & 0.654          & 17.45          & 0.737          & 0.649          & 20.98          & 0.397          & 0.649          & 18.71          \\
Ours w over compositing  &0.308 & 0.695 & 21.19 & 0.445 & 0.531 & 18.14 & 0.321 & 0.709 & 19.32 & 0.334 & 0.689 & 21.57 & 0.409 & 0.684 & 19.02          \\ 

Ours w/o warped sources & 0.250          & 0.697          & 20.89          & 0.402          & 0.534          & 18.58          & 0.263          & 0.729          & {19.31} & 0.254          & 0.704          & {22.51} & 0.322          & 0.705          & {19.65}          \\ 
Ours w/o refinement     & 0.237          & 0.675          & 20.99          & 0.380          & 0.533          & 18.28          & \textbf{0.220} & 0.717          & \textbf{20.13}          & \textbf{0.244} & 0.689          & \textbf{22.88}          & \textbf{0.239}          & \textbf{0.765}          & \textbf{21.44}          \\
Our whole pipeline      & \textbf{0.233} & \textbf{0.708} & \textbf{21.33} & \textbf{0.386} & \textbf{0.542} & \textbf{18.81} & {0.250} & \textbf{0.732} & 19.20          & {0.245} & \textbf{0.710} & 22.12          & {0.303} & {0.721} & {19.20} \\
\bottomrule
\end{tabularx}
\label{tab:abla}
\end{table*}

\subsection{Ablation study}
In this section, we conduct experiments to verify the effectiveness of each component in the proposed framework. 

\textbf{Source-view visibility estimation.}
We first remove the visibility-aware source view aggregation (indicated by Eq.~\eqref{eq:condition_on_depth}) from our framework, denoted as ``Ours w/o visibility''. 
Instead, we feed the warped source images to our refinement network directly and equally.
We expect the refinement network to learn the visibility-aware blending weights for source view images automatically. 
The results are presented in the first row of Tab.~\ref{tab:abla}. 
It can be seen that the performance drops significantly compared to our whole pipeline. 
This indicates that it is hard for the refinement network to select visible pixels from source views.

We present a visualization example of our visibility-aware aggregated result in Fig.~\ref{fig:visibility}. 
As shown in Fig.~\ref{subfig:visibility_warped_source_view}, directly warped source images contain severe ghosting artifacts due to occlusions, 
\ie, disoccluded regions in target view are filled by replicas of visible pixels from a source view. 
By using the proposed SVE module to estimate the visibility of source views, our aggregated result, Fig.~\ref{subfig:visibility_aggregated_source_view}, successfully reduces the ghosting artifacts and is much more similar to the ground truth image,  Fig.~\ref{subfig:visibility_gt}.

\textbf{Soft ray-casting.}
We first remove the soft ray-casting mechanism from our whole pipeline, expressed as ``Ours w/o ray-casting''.
Instead, we use the surface probability, \ie, the red curve in Fig.~\ref{fig:soft_ray_casting}, as the depth probability to warp and aggregate source views.
As indicated by the second row of Tab.~\ref{tab:abla}, the results are significantly inferior to our whole pipeline.
Furthermore, we replace the SRC as the conventional over alpha compositing scheme, denoted as ``Ours w over compositing''.
The results are presented in the third row of Tab.~\ref{tab:abla}.
It can be seen that our SRC is necessary and cannot be replaced by the over alpha compositing scheme.

Both SRC and over alpha compositing are neural renderers in NVS.
Over-compositing uses opacity to handle occlusions, while our method does not regress opacity for voxels.
Our input curve to SRC is obtained by majority voting from source views.
A smaller peak in the curve indicates that a surface is visible by fewer source views and a larger peak suggests that a surface is visible by more source views.
Due to the fixed weight embedding in over-compositing, the smaller peak at a nearer distance will be ignored while the larger peak will be highlighted.
By using LSTM, our SRC can be trained to decide which peak is the top-most surface.

\textbf{Refinement network.}
We further ablate the refinement network.
In doing so, we remove the warped source view images from the refinement network input, denoted as ``Ours w/o warped sources''.
As shown by the results in Tab.~\ref{tab:abla}, there is only a small performance drop compared to our whole pipeline.
This indicates that our visibility-aware aggregated images are already powerful enough to guide the refinement network to synthesize realistic images.

We further remove the refinement network from our whole pipeline, denoted as ``Ours w/o refinement''.
The results are presented in the penultimate row in Tab.~\ref{tab:abla}.
For the test scenes on the Tanks and Temples dataset, the performance drops compared to our whole pipeline.
While for the results on the DTU dataset, ``Ours w/o refinement'' achieves significantly better performance.
This is due to the huge color differences between the training (outdoor) and testing(indoor) scenes.
In practice, we suggest the users first visually measure the color differences between the training and testing scenes and then choose a suitable part of our approach.
We found that a concurrent work~\cite{riegler2020stable} provides a solution to this problem.
Interested readers are referred to this work for detailed illustration.

\textbf{Limitations.} 
The major limitation of our approach is the GPU memory. 
The required GPU memory increases with the depth plane sampling number and the input view number. 
By using a 12GB memory GPU, our approach can handle a maximum of $6$ input views and $48$ depth plane numbers. 
The advantage of inverse depth plane sampling is that it can recover near objects precisely. 
The downside is that it handles worse for scene objects with fine structures at distance, because the depth planes at distance is sampled sparsely and the correct depth of some image pixels cannot be accurately searched. 
Another limitation of our method is that we have not incorporated temporal consistency into our method when synthesizing a sequence of new views. There might be shifting pixels between the synthesized images, especially for thin objects. 
We expect these limitations can be handled in future works.

\section{Conclusion} 

In this paper, we have proposed a novel geometry-based framework for novel view synthesis. 
Different from conventional image-based rendering methods, we combine geometry estimation and image synthesis in an end-to-end framework. By doing so, inaccurately estimated geometry can be corrected by image synthesis error during training. 
Our major contribution as well as the central innovation is that we estimate the target-view depth and source-view visibility in an end-to-end self-supervised manner. 
Our network is able to generalize to unseen data without further fine-tuning. 
Experimental results demonstrate that our generated images have higher-quality than the recent state-of-the-art. 

\section{ Acknowledgments}
This research is funded in part by the ARC Centre of Excellence for Robotics Vision (CE140100016) and ARC-Discovery (DP 190102261).
% This research is supported in part by the Australian Research Council (ARC) Centre of Excellence for Robotic Vision (CE140100016),  ARC-Discovery (DP 190102261) and ARC-LIEF (190100080),  as well as a research grant from Baidu on autonomous driving. 
The first author is a China Scholarship Council (CSC)-funded PhD student to ANU. 
% We gratefully acknowledge the GPUs donated by the NVIDIA Corporation. 
We thank all anonymous reviewers and ACs for their constructive suggestions.

%------------------------------------------------------------------------

{\small
\bibliographystyle{ieee_fullname}
\bibliography{egbib}
}

\newpage
\onecolumn
\appendix
\appendixpage

\section{Robustness to Source-View Permutations} 
\begin{figure*}[ht!]
    \centering
    \subfigure[Truck]{
    \begin{minipage}{0.25\linewidth}
    \begin{sideways}
    \footnotesize
    \hspace{8mm}
    Ours
    \end{sideways}
    \includegraphics[width=0.92\linewidth]{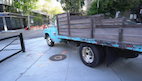}\\
    \begin{sideways}
    \footnotesize
    \hspace{1mm}
    Ours (permuted)
    \end{sideways}
    \includegraphics[width=0.92\linewidth]{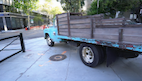}\\
    \begin{sideways}
    \footnotesize
    \hspace{2mm}
    Ground Truth
    \end{sideways}
    \includegraphics[width=0.92\linewidth]{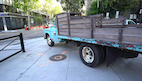}
    \end{minipage}
    % \label{subfig:EVS}
    }
    \subfigure[Train]{
    \begin{minipage}{0.23\linewidth}
    \includegraphics[width=\linewidth]{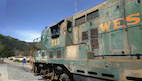}\\
    \includegraphics[width=\linewidth]{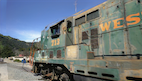}\\
    \includegraphics[width=\linewidth]{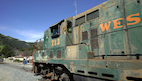}
    \end{minipage}
    % \label{subfig:FVS}
    }
    \subfigure[M60]{
    \begin{minipage}{0.23\linewidth}
    \includegraphics[width=\linewidth]{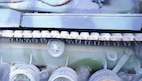}\\
    \includegraphics[width=\linewidth]{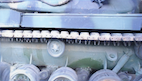}\\
    \includegraphics[width=\linewidth]{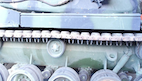}
    \end{minipage}
    % \label{subfig:ours}
    }
    \subfigure[Playground]{
    \begin{minipage}{0.23\linewidth}
    \includegraphics[width=\linewidth]{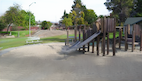}\\
    \includegraphics[width=\linewidth]{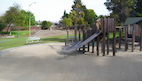}\\
    \includegraphics[width=\linewidth]{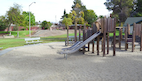}
    \end{minipage}
    % \label{subfig:GT}
    }
    \caption{ Qualitative comparison of generated images by our method with different input view orders. 
    % The first two rows present the generated images by ``Ours'' and ``Ours (permuted)'' respectively. The ground truth target images are presented in the third row. 
    The four examples are the same as those in Fig.~1 in the main paper. 
    }
    \label{fig:permutation}
\end{figure*}

In real world scenarios, the input source views are usually unordered. Thus, we take into account source-view permutation invariance when designing our framework. 
To demonstrate this, we randomly permute the input source views and feed them to the same trained model, denoted as ``Ours (permuted)''. 
Quantitative results are presented in Tab.~\ref{tab:permutation} and qualitative visualizations are provided in Fig.~\ref{fig:permutation}. 

From the results, it can be seen that there is negligible difference on the synthesized images with different input view orders. 
This demonstrates the robustness of our method to source-view permutations. 

% It can be seen the numerical evaluation results on synthesized images are almost the same with different input view orders, and the synthesized images are indistinguishable with each other. 
% This demonstrates the robustness of our method to different input view orders.

\begin{table*}[ht]
\setlength{\tabcolsep}{1pt}
\centering
\caption{Robustness of our method to different input view orders.}
\small
\begin{tabularx}{\linewidth}{l|ccc|ccc|ccc|ccc|ccc}
\toprule
\multirow{3}{*}{}       & \multicolumn{12}{c|}{Tanks and Temples}                                                                                                                                                                    & \multicolumn{3}{c}{\multirow{2}{*}{DTU}}         \\
                        & \multicolumn{3}{c|}{Truck}                        & \multicolumn{3}{c|}{Train}                        & \multicolumn{3}{c|}{M60}                          & \multicolumn{3}{c|}{Playground}                   & \multicolumn{3}{c}{}                             \\
                        & LPIPS$\downarrow$          & SSIM$\uparrow$           & PSNR$\uparrow$     & LPIPS$\downarrow$   & SSIM$\uparrow$          & PSNR$\uparrow$     & LPIPS$\downarrow$    & SSIM$\uparrow$   & PSNR$\uparrow$   & LPIPS$\downarrow$   & SSIM$\uparrow$    & PSNR$\uparrow$   & LPIPS$\downarrow$ & SSIM$\uparrow$     & PSNR$\uparrow$         \\ \midrule

% Ours w/o refinement     & 0.237          & 0.675          & 20.99          & 0.380          & 0.533          & 18.28          & \textbf{0.220} & 0.717          & 20.13          & \textbf{0.244} & 0.689          & 22.88          & 0.239          & 0.765          & 21.44         \\

Ours      & {0.233} & {0.708} & {21.33} & {0.386} & {0.542} & {18.81} & 0.250          & {0.732} & 19.20          & 0.245          & {0.710} & 22.12          & {0.177} & {0.721} & {19.20} \\
Ours (permuted) & 0.233   & 0.708  & 21.34  & 0.385   & 0.542  & 18.81  & 0.250  & 0.732  & 19.13 & 0.246    & 0.710    & 22.13    & 0.177        & 0.721       & 19.20  \\

\bottomrule
\end{tabularx}
\label{tab:permutation}
\end{table*}

\section{Residual Visualization Between Images Before and After Refinement}
\begin{figure}
    \centering
    \begin{minipage}{0.24\linewidth}
    \includegraphics[width=\linewidth]{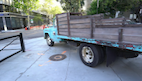}
    \includegraphics[width=\linewidth]{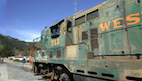}
    \includegraphics[width=\linewidth]{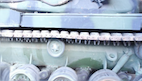}
    \includegraphics[width=\linewidth]{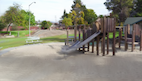}
    \centerline{Aggregated Images}
    \end{minipage}
    \begin{minipage}{0.24\linewidth}
    \includegraphics[width=\linewidth]{results_Truck_ours_24.png}
    \includegraphics[width=\linewidth]{results_Train_ours_11.png}
    \includegraphics[width=\linewidth]{results_M60_ours_16.png}
    \includegraphics[width=\linewidth]{results_Playground_ours_12.png}
    \centerline{Output Images}
    \end{minipage}
    \begin{minipage}{0.24\linewidth}
    \includegraphics[width=\linewidth]{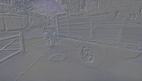}
    \includegraphics[width=\linewidth]{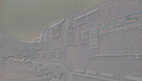}
    \includegraphics[width=\linewidth]{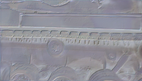}
    \includegraphics[width=\linewidth]{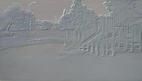}
    \centerline{Residual Images}
    \end{minipage}
    \begin{minipage}{0.24\linewidth}
    \includegraphics[width=\linewidth]{results_Truck_tgt_24.png}
    \includegraphics[width=\linewidth]{results_Train_tgt_11.png}
    \includegraphics[width=\linewidth]{results_M60_tgt_16.png}
    \includegraphics[width=\linewidth]{results_Playground_tgt_12.png}
    \centerline{Ground Truths}
    \end{minipage}
    \caption{Visualization of aggregated images, output images after refinement, residual images between the aggregated images and the output images, and the ground truths. The four examples are from scenes ``\textit{Truck}'', ``\textit{Train}'', ``\textit{M60}'' and ``\textit{Playground}'', respectively. Their corresponding input images are presented in Fig.\ref{fig:Truck}, Fig.\ref{fig:Train}, Fig.\ref{fig:M60}, and Fig.\ref{fig:Playground}, respectively. }
    \label{fig:residual}
\end{figure}

We visualize the aggregated images by Eq.(2) in the paper and the corresponding output images after the refinement network in Fig.~\ref{fig:residual}. 
To show the differences, we also present the residual images between the aggregated images and final output images. 

\section{Additional Visualization on the DTU Dataset}
\begin{figure*}[ht!]
    \centering
    \subfigure[EVS~\cite{choi2019extreme}]{
    \begin{minipage}{0.235\linewidth}
    \includegraphics[width=\linewidth]{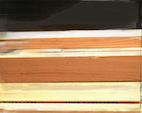}\\
    \includegraphics[width=\linewidth]{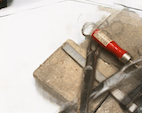}\\
    \includegraphics[width=\linewidth]{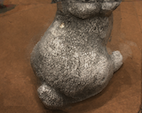}\\
    \includegraphics[width=\linewidth]{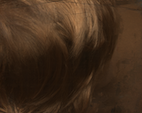}\\
    \includegraphics[width=\linewidth]{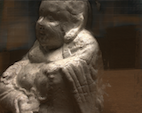}\\
    \end{minipage}
    }
    \subfigure[FVS~\cite{riegler2020free}]{
    \begin{minipage}{0.235\linewidth}
    \includegraphics[width=\linewidth]{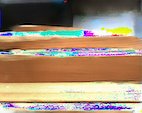}\\
    \includegraphics[width=\linewidth]{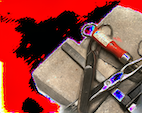}\\
    \includegraphics[width=\linewidth]{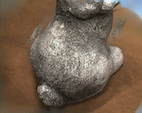}\\
    \includegraphics[width=\linewidth]{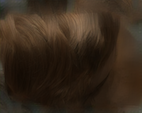}\\
    \includegraphics[width=\linewidth]{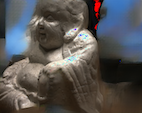}\\
    \end{minipage}
    }
    \subfigure[Ours]{
    \begin{minipage}{0.235\linewidth}
   \includegraphics[width=\linewidth]{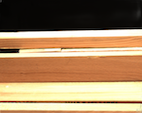}\\
    \includegraphics[width=\linewidth]{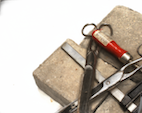}\\
    \includegraphics[width=\linewidth]{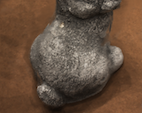}\\
    \includegraphics[width=\linewidth]{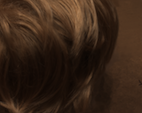}\\
    \includegraphics[width=\linewidth]{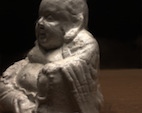}\\
    \end{minipage}
    }
    \subfigure[Ground Truth]{
    \begin{minipage}{0.235\linewidth}
    \includegraphics[width=\linewidth]{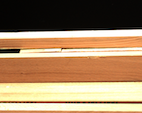}\\
    \includegraphics[width=\linewidth]{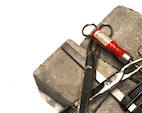}\\
    \includegraphics[width=\linewidth]{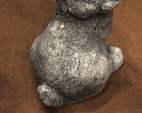}\\
    \includegraphics[width=\linewidth]{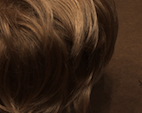}\\
    \includegraphics[width=\linewidth]{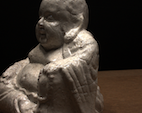}\\
    \end{minipage}
    }
    \caption{ Additional qualitative visualization of generated results on the DTU dataset with six views as input.
    }
    \label{fig:stoa_dtu_addi}
\end{figure*}

% Due to the space limit, we provide two visualization examples on the DTU dataset in the main paper. 
In this section, we present more visualization examples on the DTU dataset, as shown in Fig.~\ref{fig:stoa_dtu_addi}. 
% Qualitative comparisons with the recent state-of-the-art are presented in Fig.~\ref{}. 
Scenes in the DTU dataset are simpler than those in the Tanks and Temples dataset, \ie, there is always a single salient object in each image. 
However, there are large textureless regions and some of the object surfaces (\eg, the second example in Fig.~\ref{fig:stoa_dtu_addi}) are reflective.

For those regions, COLMAP almost fails to predict the depths, and thus the images synthesized by FVS~\cite{riegler2020free} suffer severe artifacts. 
EVS estimates target-view depths and source-view visibility from the source view depths. When the source view depths are inaccurate and/or disagree with each other, the error will be accumulated to the final synthesized image. 
In contrast, our method does not rely on the accuracy of source view depths and is able to handle textureless and reflective regions. 
Thus our synthesized images are much closer to ground-truth images. 

% Although the DTU dataset is a more constrained dataset than the Tanks and Temples~\cite{riegler2020free}, it contains large textureless regions.  

\begin{figure}[ht!]
    \centering
    \begin{minipage}{0.45\linewidth}
    \begin{minipage}{\linewidth}
    \subfigure[Input Source-View Images]{
    \begin{minipage}{0.45\linewidth}
    \vspace{-0.5mm}
    \includegraphics[width=\linewidth]{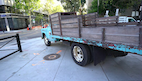}\\
    \centerline{\scriptsize $\text{Rot}=[0.90^{\circ}, 1.73^{\circ}, -0.71^{\circ}]$}
    \centerline{\scriptsize $\text{Trans}=[-25.56\text{cm}, -2.50\text{cm}, 3.32\text{cm}]$}
    \vspace{-0.5mm}
    \includegraphics[width=\linewidth]{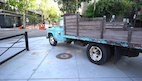}\\
    \centerline{\scriptsize $\text{Rot}=[-0.95^{\circ}, -4.10^{\circ}, 1.67^{\circ}]$}
    \centerline{\scriptsize $\text{Trans}=[24.88\text{cm}, 0.04\text{cm}, 22.57\text{cm}]$}
    \vspace{-0.5mm}
    \includegraphics[width=\linewidth]{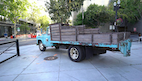}\\
    \centerline{\scriptsize $\text{Rot}=[-6.46^{\circ}, -8.52^{\circ}, -2.26^{\circ}]$}
    \centerline{\scriptsize $\text{Trans}=[-51.16\text{cm}, 22.23\text{cm}, 100.32\text{cm}]$}
    \vspace{-0.5mm}
    \includegraphics[width=\linewidth]{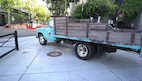}\\
    \centerline{\scriptsize $\text{Rot}=[-1.34^{\circ}, -11.69^{\circ}, 1.28^{\circ}]$}
    \centerline{\scriptsize $\text{Trans}=[-23.63\text{cm}, -3.57\text{cm}, 57.15\text{cm}]$}
    \vspace{-0.5mm}
    \includegraphics[width=\linewidth]{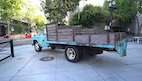}\\
    \centerline{\scriptsize $\text{Rot}=[-6.45^{\circ}, -11.41^{\circ}, -2.27^{\circ}]$}
    \centerline{\scriptsize $\text{Trans}=[-68.72\text{cm}, 23.02\text{cm}, 93.47\text{cm}]$}
    \vspace{-0.5mm}
    \includegraphics[width=\linewidth]{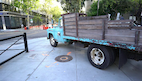}\\
    \centerline{\scriptsize $\text{Rot}=[-0.41^{\circ}, -1.08^{\circ}, 0.94^{\circ}]$}
    \centerline{\scriptsize $\text{Trans}=[24.33\text{cm}, 30.47\text{cm}, -4.96\text{cm}]$}\\
    \end{minipage}
    }
    \subfigure[Warped Source-View Images]{
    \begin{minipage}{0.45\linewidth}
    \includegraphics[width=\linewidth]{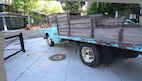}
    \\ \\ \\ 
    \includegraphics[width=\linewidth]{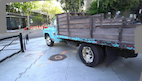}
    \\ \\ \\ 
    \includegraphics[width=\linewidth]{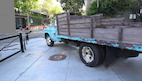}
    \\ \\ \\ 
    \includegraphics[width=\linewidth]{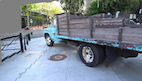}
    \\ \\ \\ 
    \includegraphics[width=\linewidth]{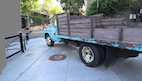}
    \\ \\ \\ 
    \includegraphics[width=\linewidth]{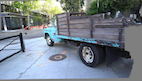}
    \\ \\ \\ 
    \end{minipage}
    }
    \subfigure[Target-View Ground Truth]{
    \begin{minipage}{0.45\linewidth}
    \includegraphics[width=\linewidth]{results_Truck_tgt_24.png}
    \end{minipage}
    }
    \hfill
    \subfigure[Aggregated Image]{
    \begin{minipage}{0.45\linewidth}
    \includegraphics[width=\linewidth]{results_Truck_aggregated_24.png}
    \end{minipage}
    }\\
    \end{minipage}
    \caption{Qualitative illustration on our visibility-aware aggregation (\textit{Truck}). }
    \label{fig:Truck}
    \end{minipage}
    \hfill
    \begin{minipage}{0.45\linewidth}
    \begin{minipage}{\linewidth}
    \subfigure[Input Source-View Images]{
    \begin{minipage}{0.45\linewidth}
    \vspace{-0.5mm}
    \includegraphics[width=\linewidth]{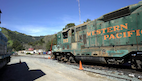}\\
    \centerline{\scriptsize $\text{Rot}=[11.73^{\circ}, 5.42^{\circ}, 9.12^{\circ}]$}
    \centerline{\scriptsize $\text{Trans}=[7.70\text{cm}, 1.27\text{cm}, 175.36\text{cm}]$}
    \vspace{-0.5mm}
    \includegraphics[width=\linewidth]{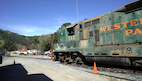}\\
    \centerline{\scriptsize $\text{Rot}=[16.40^{\circ}, 1.85^{\circ}, 14.23^{\circ}]$}
    \centerline{\scriptsize $\text{Trans}=[42.91\text{cm}, -8.60\text{cm}, 121.41\text{cm}]$}
    \vspace{-0.5mm}
    \includegraphics[width=\linewidth]{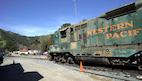}\\
    \centerline{\scriptsize $\text{Rot}=[15.83^{\circ}, 2.58^{\circ}, 13.22^{\circ}]$}
    \centerline{\scriptsize $\text{Trans}=[8.29\text{cm}, -1.48\text{cm}, 157\text{cm}]$}
    \vspace{-0.5mm}
    \includegraphics[width=\linewidth]{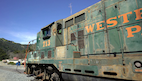}\\
    \centerline{\scriptsize $\text{Rot}=[0.10^{\circ}, -2.71^{\circ}, 0.55^{\circ}]$}
    \centerline{\scriptsize $\text{Trans}=[-21.16\text{cm}, -0.34\text{cm}, 17.46\text{cm}]$}
    \vspace{-0.5mm}
    \includegraphics[width=\linewidth]{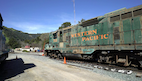}\\
    \centerline{\scriptsize $\text{Rot}=[20.18^{\circ}, 4.12^{\circ}, 18.33^{\circ}]$}
    \centerline{\scriptsize $\text{Trans}=[-52.76\text{cm}, 5.25\text{cm}, 260.19\text{cm}]$}
    \vspace{-0.5mm}
    \includegraphics[width=\linewidth]{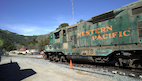}\\
    \centerline{\scriptsize $\text{Rot}=[18.20^{\circ}, 1.54^{\circ}, 15.74^{\circ}]$}
    \centerline{\scriptsize $\text{Trans}=[-24.79\text{cm}, 2.41\text{cm}, 198.15\text{cm}]$}\\
    \end{minipage}
    }
    \subfigure[Warped Source-View Images]{
    \begin{minipage}{0.45\linewidth}
    \includegraphics[width=\linewidth]{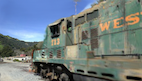}
    \\ \\ \\ 
    \includegraphics[width=\linewidth]{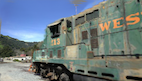}
    \\ \\ \\ 
    \includegraphics[width=\linewidth]{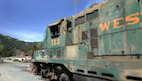}
    \\ \\ \\ 
    \includegraphics[width=\linewidth]{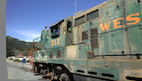}
    \\ \\ \\ 
    \includegraphics[width=\linewidth]{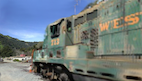}
    \\ \\ \\ 
    \includegraphics[width=\linewidth]{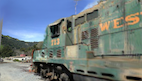}
    \\ \\ \\ 
    \end{minipage}
    }
    \subfigure[Target-View Ground Truth]{
    \begin{minipage}{0.45\linewidth}
    \includegraphics[width=\linewidth]{results_Train_tgt_11.png}
    \end{minipage}
    }
    \hfill
    \subfigure[Aggregated Image]{
    \begin{minipage}{0.45\linewidth}
    \includegraphics[width=\linewidth]{results_Train_aggregated_11.png}
    \end{minipage}
    }\\
    \end{minipage}
    \caption{Qualitative illustration on our visibility-aware aggregation (\textit{Train}). 
    % This figure corresponds to the second example in Fig.~\ref{fig:stoa_tat} in the main paper.
    }
    \label{fig:Train}
    \end{minipage}
\end{figure}

\begin{figure}[ht!]
    \centering
    \begin{minipage}{0.45\linewidth}
    \begin{minipage}{\linewidth}
    \subfigure[Input Source-View Images]{
    \begin{minipage}{0.45\linewidth}
    \vspace{-0.5mm}
    \includegraphics[width=\linewidth]{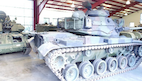}\\
    \centerline{\scriptsize $\text{Rot}=[137.92^{\circ}, 5.13^{\circ}, -139.39^{\circ}]$}
    \centerline{\scriptsize $\text{Trans}=[56.66\text{cm}, 16.41\text{cm}, 211.42\text{cm}]$}
    \vspace{-0.5mm}
    \includegraphics[width=\linewidth]{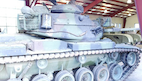}\\
    \centerline{\scriptsize $\text{Rot}=[137.13^{\circ}, 1.47^{\circ}, -138.50^{\circ}]$}
    \centerline{\scriptsize $\text{Trans}=[-36.27\text{cm}, 28.48\text{cm}, 120.76\text{cm}]$}
    \vspace{-0.5mm}
    \includegraphics[width=\linewidth]{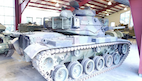}\\
    \centerline{\scriptsize $\text{Rot}=[142.98^{\circ}, 13.25^{\circ}, -144.97^{\circ}]$}
    \centerline{\scriptsize $\text{Trans}=[0.29\text{cm}, 15.12\text{cm}, 210.44\text{cm}]$}
    \vspace{-0.5mm}
    \includegraphics[width=\linewidth]{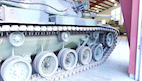}\\
    \centerline{\scriptsize $\text{Rot}=[145.35^{\circ}, 16.37^{\circ}, -147.98^{\circ}]$}
    \centerline{\scriptsize $\text{Trans}=[-106.24\text{cm}, -14.62\text{cm}, 73.63\text{cm}]$}
    \vspace{-0.5mm}
    \includegraphics[width=\linewidth]{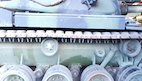}\\
    \centerline{\scriptsize $\text{Rot}=[28.33^{\circ}, -8.87^{\circ}, -28.85^{\circ}]$}
    \centerline{\scriptsize $\text{Trans}=[-33.01\text{cm}, -1.91\text{cm}, 8.70\text{cm}]$}
    \vspace{-0.5mm}
    \includegraphics[width=\linewidth]{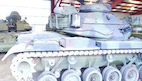}\\
    \centerline{\scriptsize $\text{Rot}=[109.67^{\circ}, -7.22^{\circ}, -110.39^{\circ}]$}
    \centerline{\scriptsize $\text{Trans}=[53.16\text{cm}, 23.28\text{cm}, 128.66\text{cm}]$}\\
    \end{minipage}
    }
    \subfigure[Warped Source-View Images]{
    \begin{minipage}{0.45\linewidth}
    \includegraphics[width=\linewidth]{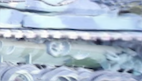}
    \\ \\ \\ 
    \includegraphics[width=\linewidth]{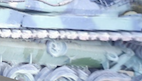}
    \\ \\ \\ 
    \includegraphics[width=\linewidth]{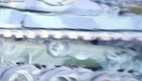}
    \\ \\ \\ 
    \includegraphics[width=\linewidth]{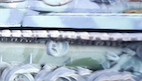}
    \\ \\ \\ 
    \includegraphics[width=\linewidth]{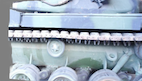}
    \\ \\ \\ 
    \includegraphics[width=\linewidth]{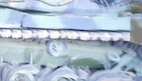}
    \\ \\ \\ 
    \end{minipage}
    }
    \subfigure[Target-View Ground Truth]{
    \begin{minipage}{0.45\linewidth}
    \includegraphics[width=\linewidth]{results_M60_tgt_16.png}
    \end{minipage}
    }
    \hfill
    \subfigure[Aggregated Image]{
    \begin{minipage}{0.45\linewidth}
    \includegraphics[width=\linewidth]{results_M60_aggregated_16.png}
    \end{minipage}
    }\\
    \end{minipage}
    \caption{Qualitative illustration on our visibility-aware aggregation (\textit{M60}). 
    % This figure corresponds to the second example in Fig.~\ref{fig:stoa_tat} in the main paper.
    }
    \label{fig:M60}
    \end{minipage}
    \hfill
    \begin{minipage}{0.45\linewidth}
    \begin{minipage}{\linewidth}
    \subfigure[Input Source-View Images]{
    \begin{minipage}{0.45\linewidth}
    \vspace{-0.5mm}
    \includegraphics[width=\linewidth]{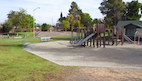}\\
    \centerline{\scriptsize $\text{Rot}=[-0.82^{\circ}, -2.05^{\circ}, 0
    34^{\circ}]$}
    \centerline{\scriptsize $\text{Trans}=[18.11\text{cm}, -7.50\text{cm}, 168.89\text{cm}]$}
    \vspace{-0.5mm}
    \includegraphics[width=\linewidth]{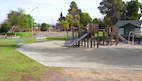}\\
    \centerline{\scriptsize $\text{Rot}=[-0.68^{\circ}, -1.72^{\circ}, 0.21^{\circ}]$}
    \centerline{\scriptsize $\text{Trans}=[-2.12\text{cm}, -7.24\text{cm}, 167.73\text{cm}]$}
    \vspace{-0.5mm}
    \includegraphics[width=\linewidth]{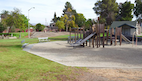}\\
    \centerline{\scriptsize $\text{Rot}=[-0.83^{\circ}, -4.84^{\circ}, 0.17^{\circ}]$}
    \centerline{\scriptsize $\text{Trans}=[15.00\text{cm}, -6.87\text{cm}, 171.10\text{cm}]$}
    \vspace{-0.5mm}
    \includegraphics[width=\linewidth]{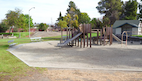}\\
    \centerline{\scriptsize $\text{Rot}=[-0.62^{\circ}, -1.74^{\circ}, 0.08^{\circ}]$}
    \centerline{\scriptsize $\text{Trans}=[-25.01\text{cm}, -7.28\text{cm}, 166.32\text{cm}]$}
    \vspace{-0.5mm}
    \includegraphics[width=\linewidth]{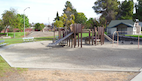}\\
    \centerline{\scriptsize $\text{Rot}=[-0.55^{\circ}, -2.53^{\circ}, -0.09^{\circ}]$}
    \centerline{\scriptsize $\text{Trans}=[-54.80\text{cm}, -7.45\text{cm}, 165.68\text{cm}]$}
    \vspace{-0.5mm}
    \includegraphics[width=\linewidth]{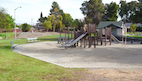}\\
    \centerline{\scriptsize $\text{Rot}=[-0.69^{\circ}, -10.34^{\circ}, -0.33^{\circ}]$}
    \centerline{\scriptsize $\text{Trans}=[-15.70\text{cm}, -6.60\text{cm}, 173.51\text{cm}]$}\\
    \end{minipage}
    }
    \subfigure[Warped Source-View Images]{
    \begin{minipage}{0.45\linewidth}
    \includegraphics[width=\linewidth]{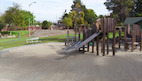}
    \\ \\ \\ 
    \includegraphics[width=\linewidth]{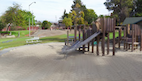}
    \\ \\ \\ 
    \includegraphics[width=\linewidth]{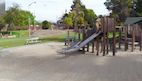}
    \\ \\ \\ 
    \includegraphics[width=\linewidth]{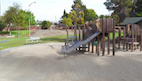}
    \\ \\ \\ 
    \includegraphics[width=\linewidth]{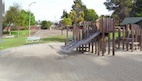}
    \\ \\ \\ 
    \includegraphics[width=\linewidth]{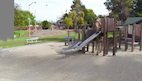}
    \\ \\ \\ 
    \end{minipage}
    }
    \subfigure[Target-View Ground Truth]{
    \begin{minipage}{0.45\linewidth}
    \includegraphics[width=\linewidth]{results_Playground_tgt_12.png}
    \end{minipage}
    }
    \hfill
    \subfigure[Aggregated Image]{
    \begin{minipage}{0.45\linewidth}
    \includegraphics[width=\linewidth]{results_Playground_aggregated_12.png}
    \end{minipage}
    }\\
    \end{minipage}
    \caption{Qualitative illustration on our visibility-aware aggregation (\textit{Playground}). 
    % This figure corresponds to the second example in Fig.~\ref{fig:stoa_tat} in the main paper.
    }
    \label{fig:Playground}
    \end{minipage}
\end{figure}

\section{Additional Visualization on the Source-View Visibility Estimation (SVE)}

In this section, we provide more visualization results on our visibility-aware aggregation mechanism, which is indicated by Eq.~(2)
% \eqref{eq:aggregation} 
in the main paper. 
% Since we also want to provide the source images and relative camera movements between source and target view cameras
We choose the four examples presented in Fig.~1 in the main paper for this visualization. %, and 
% In addition, in order to provide source images and relative camera movements between source and target view cameras, we choose the four examples in Fig.~1 in the main paper for this visualization. 
The results are illustrated in Fig.~\ref{fig:Truck}, Fig.~\ref{fig:Train}, Fig.~\ref{fig:M60} and Fig.~\ref{fig:Playground}, respectively. 
The camera movements between source and target views are provided under each input source view image in the Figures. 

% For camera movements between source and target views in the DTU dataset, please refer to our supplementary video for a perceptual understanding.

\section{Qualitative Illustration on the Proposed Soft Ray-Casting (SRC) }
The proposed soft ray-casting (SRC) mechanism is one of the key components in our framework. 
It converts a multi-modal surface existence probability along a viewing ray to a single-modal depth probability. 

\begin{figure*}
    \centering
    \subfigure[Generated depth (the first row) and image (the second row) by ``Ours w/o ray-casting''. ]{
    \begin{minipage}{0.23\linewidth}
    \includegraphics[width=\linewidth]{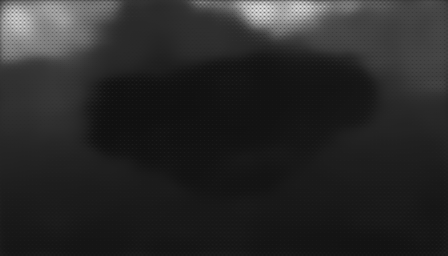}
    \includegraphics[width=\linewidth]{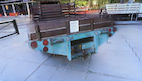}
    \centerline{\footnotesize Truck}
    \end{minipage}
    \begin{minipage}{0.23\linewidth}
    \includegraphics[width=\linewidth]{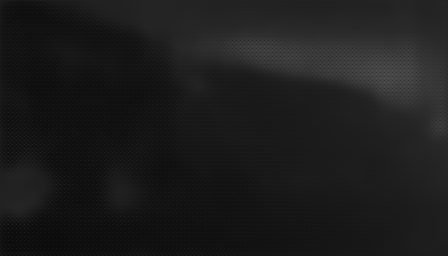}
    \includegraphics[width=\linewidth]{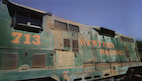}
    \centerline{\footnotesize Train}
    \end{minipage}
    \begin{minipage}{0.23\linewidth}
    \includegraphics[width=\linewidth]{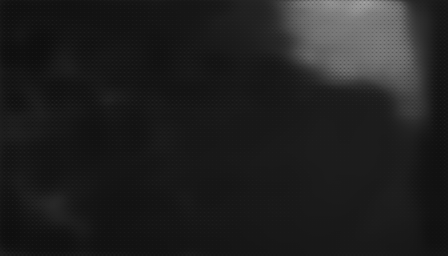}
    \includegraphics[width=\linewidth]{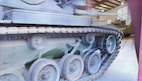}
    \centerline{\footnotesize M60}
    \end{minipage}
    \begin{minipage}{0.23\linewidth}
    \includegraphics[width=\linewidth]{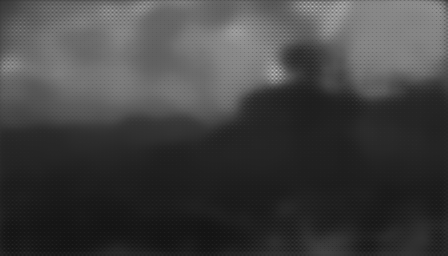}
    \includegraphics[width=\linewidth]{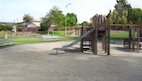}
    \centerline{\footnotesize ground}
    \end{minipage}
    }
    \subfigure[Generated depth (the first row) and image (the second row) by ``Our whole pipeline''. ]{
    \begin{minipage}{0.23\linewidth}
    \includegraphics[width=\linewidth]{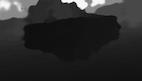}
    \includegraphics[width=\linewidth]{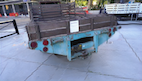}
    \centerline{\footnotesize Truck}
    \end{minipage}
    \begin{minipage}{0.23\linewidth}
    \includegraphics[width=\linewidth]{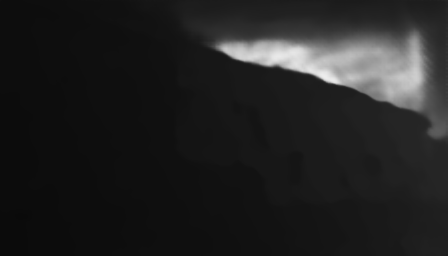}
    \includegraphics[width=\linewidth]{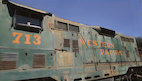}
    \centerline{\footnotesize Train}
    \end{minipage}
    \begin{minipage}{0.23\linewidth}
    \includegraphics[width=\linewidth]{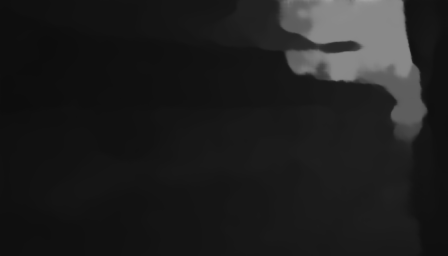}
    \includegraphics[width=\linewidth]{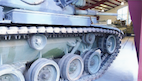}
    \centerline{\footnotesize M60}
    \end{minipage}
    \begin{minipage}{0.23\linewidth}
    \includegraphics[width=\linewidth]{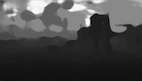}
    \includegraphics[width=\linewidth]{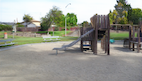}
    \centerline{\footnotesize ground}
    \end{minipage}
    }
    \subfigure[Ground Truth]{
    \begin{minipage}{0.23\linewidth}
    \includegraphics[width=\linewidth]{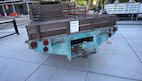}
    \centerline{\footnotesize Truck}
    \end{minipage}
    \begin{minipage}{0.23\linewidth}
    \includegraphics[width=\linewidth]{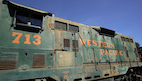}
    \centerline{\footnotesize Train}
    \end{minipage}
    \begin{minipage}{0.23\linewidth}
    \includegraphics[width=\linewidth]{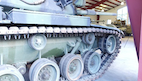}
    \centerline{\footnotesize M60}
    \end{minipage}
    \begin{minipage}{0.23\linewidth}
    \includegraphics[width=\linewidth]{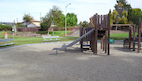}
    \centerline{\footnotesize ground}
    \end{minipage}
    }
    \caption{Qualitative comparison between ``Ours w/o ray-casting'' and ``Our whole pipeline''. 
    % The first two rows are generated depth and images by ``Ours w_0 ray-casting'' respectively
    % First row: generated depth by ``Ours w/o ray-casting''; Scond row: generated image by ``Ours w/o ray-casting''; Third row: generated depth by ``Ours''; Forth row: generated image by ``Ours''; Lat row: ground truth. 
    }
    \label{fig:depth}
\end{figure*}

We demonstrate its necessity by removing it from our whole pipeline, denoted as ``Ours w/o ray-casting'' in the main paper. 
Fig.~\ref{fig:depth} presents the qualitative comparison. For the depth visualization, we apply a softargmax to the surface probability distribution (in ``Ours w/o ray-casting'') or the depth probability distribution (in ``Our whole pipeline'').  
From Fig.~\ref{fig:depth}, it can be observed that the generated depth and image by ``Ours w/o ray casting'' are more blurred than ``Our whole pipeline''.
This is mainly because the surface probability distribution is always multi-modal and it does not reflect pure depth information, 
demonstrating the effectiveness of the proposed soft ray-casting mechanism.

\end{document}